\documentclass[11pt]{article} % For LaTeX2e
\usepackage{my_style,times}

\usepackage{amsmath,amsfonts,bm}

\def\eqref#1{equation~\ref{#1}}
\def\1{\bm{1}}

\DeclareMathAlphabet{\mathsfit}{\encodingdefault}{\sfdefault}{m}{sl}
\SetMathAlphabet{\mathsfit}{bold}{\encodingdefault}{\sfdefault}{bx}{n}

\usepackage[utf8]{inputenc} % allow utf-8 input
\usepackage[T1]{fontenc}    % use 8-bit T1 fonts
\usepackage{url}            % simple URL typesetting

\usepackage{booktabs}       % professional-quality tables
\usepackage[round]{natbib}
\usepackage{amsfonts}       % blackboard math symbols
\usepackage{nicefrac}       % compact symbols for 1/2, etc.
\usepackage{microtype}      % microtypography
\usepackage{amsmath,bm,dsfont}
\usepackage{comment}
\usepackage{algorithm,algpseudocode}
\usepackage{multirow,tabulary,graphicx,subcaption,wrapfig}
\usepackage[font={small}]{caption}
\usepackage{xpatch}

\usepackage[colorlinks,pdfpagelabels,plainpages=false]{hyperref}
\usepackage{xcolor}
\definecolor{dark-red}{rgb}{0.4,0.15,0.15}
\definecolor{dark-blue}{rgb}{0.15,0.15,0.4}
\definecolor{medium-blue}{rgb}{0,0,0.5}
\hypersetup{
   colorlinks, linkcolor={dark-blue},
   citecolor={dark-blue}, urlcolor={medium-blue}
}

\newcolumntype{K}[1]{>{\centering\arraybackslash}m{#1}}

\title{\bf AutoLoss: Learning Discrete Schedules for Alternate Optimization}

\author{
  Haowen Xu$^{\dagger}$, Hao Zhang$^{\dagger}$, Zhiting Hu, Xiaodan Liang, \\
  Ruslan Salakhutdinov, Eric Xing
  \and 
  {\small  Carnegie Mellon University and Petuum Inc.}
  \\
  {\small $^{\dagger}$ equal contributions}
}

\begin{document}
\date{}

\maketitle

\begin{abstract}
\noindent
Many machine learning problems involve iteratively and alternately optimizing different task objectives with respect to different sets of parameters. Appropriately scheduling the optimization of a task objective or a set of parameters is usually crucial to the quality of convergence. In this paper, we present \emph{AutoLoss}, a meta-learning framework that automatically learns and determines the optimization schedule. AutoLoss provides a generic way to represent and learn the discrete optimization schedule from metadata, allows for a dynamic and data-driven schedule in ML problems that involve alternating updates of different parameters or from different loss objectives.
We apply AutoLoss on four ML tasks: $d$-ary quadratic regression, classification using a multi-layer perceptron (MLP), image generation using GANs, and multi-task neural machine translation (NMT). We show that the AutoLoss controller is able to capture the distribution of better optimization schedules that result in higher quality of convergence on all four tasks. The trained AutoLoss controller is generalizable -- it can guide  and improve the learning of a new task model with different specifications, or on different datasets.
\end{abstract}

\section{Introduction}
Many machine learning (ML) problems involve iterative alternate optimization of different objectives $\{\ell_m\}_{m=1}^M$ w.r.t different sets of parameters $\{\bm{\theta}_n\}_{n=1}^N$ until a global consensus is reached. For instances, in training generative adversarial networks (GANs)~\citep{goodfellow2014generative}, parameters of the generator and the discriminator are alternately updated to an equilibrium; 
in many multi-task learning problems~\citep{argyriou2007multi}, one usually has to alternate the optimization of different task-specific objectives on corresponded data, until the target task performance is maximized.
%in a typical expectation-maximization (EM) framework~\citep{moon1996expectation}, one usually has to alternate variable steps of computation to obtain an estimation of the expectation, then variable optimization steps for model parameters, until the objective (e.g. likelihood) is maximized. 
In these processes, one needs to determine which objective $\ell_m$ and which set of parameters $\bm{\theta}_n$ to choose at each step, and subsequently, how many iteration steps to perform for the subproblem $\min_{\bm{\theta}_n} \ell_m$. We refer to this as determining an \emph{optimization schedule} (or update schedule).
%How to optimize the loss is still a problem. however, the optimization procedures affxect the quality of the solution. Take another examples.

While extensive research has been focused on developing better optimization algorithms or update rules~\citep{kingma2014adam,bello2017neural,duchi2011adaptive,sutskever2013importance}, how to select optimization schedules has remained less studied. 
%, the other problem ``when to solve which task'' has remained unclear in many ML applications. To see this, consider the training of GANs, the update for generator and for discriminator needs to be carefully balanced in case one 
%Our motivation comes from the common practice of handling this iterative and alternative optimization process: one usually has to design an appropriate \emph{update schedule}, determining which subset of loss terms $\mathcal{L}_{t}$ and subset of parameters $\bm{\Theta}_{t}$ to choose at step $t$, and once chose, how many iterative steps shall be performed for the subproblem $\min_{\bm{\Theta}_{t}} \mathcal{L}_{t}$. 
When the objective is complex (e.g. non-convex or combinatorial) and the parameters to be optimized are high-dimensional, the optimization schedule can directly impact the quality of convergence. However, we hypothesize that the schedule is learnable in a data-driven way, with the following empirical evidence: %(while we defer a theoretical justification of the hypothesis to the future work): 
(1) The optimization of many ML models is sensitive to the update schedule. For examples, the updates of the generator and the discriminator in GANs are carefully reconciled to avoid otherwise model collapse or gradient vanishing~\citep{goodfellow2014generative,radford2015unsupervised}; 
%In EM with stochastic variational inference (SVI)~\citep{hoffman2013stochastic}, different E-steps of variational updates and M-steps of parameter updates may result in different efficiency or quality of convergence~\citep{welling2011bayesian, blei2017variational}, etc. 
In solving many multi-task learning or regularizer-augmented objectives, the optimization target $\mathcal{L}$ is a combination of multiple task-specific objectives. It is desirable to weight each objective differently as $\mathcal{L} = \sum_{m=1}^M \lambda_m \ell_m$, while different values of $\{\lambda_m\}_{m=1}^M$ result in different (local) optima.
%Though the combined loss $\mathcal{L}$ is optimized directly w.r.t. the whole set of parameters $\bm{\Theta}$ (therefore inequivalent with our setting where a subset of $\mathcal{L}$ are optimized over a subset of $\mathcal{\Theta}$ at each step), 
This indicates that different loss terms shall not be treated equally, and achieving the best downstream task performance requires optimizing every $\ell_m$ to different extents. (2) Previous research and practice have suggested that there do exist optimization schedules that are more probable to produce better convergence than random ones, e.g. ~\citet{arjovsky2017wasserstein} and~\citet{salimans2016improved} suggest that keeping the steps of updating the generator and discriminator of GANs at $K:1 (K > 1)$ leads to faster and more stable training of GANs. %In a L1 reguarlized objective, a better value of $\lambda$ for the L1 loss term can usually be found via manual grid searching, model section~\citep{}, or Bayesian optimization.

%formulation
Based on the hypothesis, in this paper, we develop \emph{AutoLoss}, a generic meta-learning framework to automatically determine the optimization schedule in iterative and alternate optimization processes. AutoLoss introduces a parametric controller attached to an alternate optimization task. The controller is trained to capture the relations between the past history and the current state of the optimization process, and the next step of the decision on the update schedule. It takes as input a set of status features, and decides which objectives from $\{\ell_m\}_{m=1}^M$ to optimize, and which set of parameters from $\{\bm{\theta}_n\}_{n=1}^N$ to update. The controller is trained via policy gradient to maximize the eventual outcome of the optimization (e.g. downstream task performance). Once trained, it can guide the optimization of task models %by providing scheduling information: it first predicts a set of losses and parameters, and instructs the task model to only perform one step of optimization over them, and eventually helps the task model 
to achieve higher quality of convergence faster, by predicting better schedules.

To evaluate the effectiveness of AutoLoss, we instantiate it on four typical ML tasks: $d$-ary quadratic regression, classification using a multi-layer perceptron (MLP), image generation using GANs, and neural machine translation (NMT) based on multi-task learning. We propose an effective set of features and reward functions that are suitable for the controllers' learning and decisions. We show that, on all four tasks, the AutoLoss controller is able to capture the distribution of better optimization schedules that result in higher quality of convergence on the corresponding task than strong baselines. For examples, on quadratic regression with L1 regularization, it learns to detect the potential risk of overfitting, and incorporates L1 regularization when necessary, helps the task model converge to better results that can hardly be achieved by optimizing linear combinations of objective terms. on GANs, the AutoLoss controller learns to balance the training of generator and discriminator dynamically, and report both faster per-epoch convergence and better quality of generators after convergence, compared to fixed heuristic-driven schedules. On machine translation, it automatically learns to resemble human-tuned update schedules while being more flexible, and reports better perplexity results. %We further show that the learned controller on a specific task can usually be transferred to help another similar task.

% how to solve subtasks have been extensively developed, but how to develop the ``update schedule'' is critical but less studied (quality-determined). e.g. loss minimization where the objective is a linear combination of terms, or in GAN/stochastic MCMC, how many steps to learn is determined by heurastics. 
%Existing methods on choosing how to optimize the loss: grid search, hyperparameter tuning, Bayesian methods, heuristics.
%While existing approaches to finding a better schedule are usually by simple heuristics and are manually expensive, we develop a generic AutoLoss framework to address this problem. 

%In this work, we propose a framework to automatically learn to iterative over losses. %unlock the possibility of using a dynamically data-driven schedule instead of a static and fixed one.

%Briefly describe the workflow of our framework
%For a specific alternative optimization task, AutoLoss is trying to learn a parametrized meta-model to capture the distribution of update schedules that yields better performance and downstream tasks. The meta model makes decision at each optimization step to help the optimizer choose which terms and which set of parameters to optimize, and the learned meta model on a specific task can usually be transferred to help another similar task.

In summary, we make the following contributions in this paper: (1) We present a unified formulation for iterative and alternate optimization processes, based on which, we develop AutoLoss, a generic framework to learn the discrete optimization schedule of such processes using reinforcement learning (RL). To our knowledge, this is the first framework that tries to learn the optimization schedule in a data-driven way.
(2) We instantiate AutoLoss on four ML tasks: $d$-ary regression, MLP classification, GANs, and NMT. We propose a novel set of features and reward functions to facilitate the training of AutoLoss controllers.
(3) We empirically demonstrate AutoLoss' efficacy: it delivers higher quality of convergence for all four tasks on synthetic and real dataset than strong baselines. Training AutoLoss controller has acceptable overhead less than most hyperparameter searching methods; the trained AutoLoss controller is generalizable -- it can guide and improve the training of a new task model with different specifications, or on different dataset.

% summarize the contributions.
\section{Related Work}
\noindent \textbf{Alternate Optimization.} 
Many ML models are trained using algorithms with iterative and alternate workflows, such as EM~\citep{moon1996expectation}, stochastic gradient descent (SGD)~\citep{bottou2010large}, coordinate descent~\citep{wright2015coordinate}, multi-task learning~\citep{zhang2017survey}, etc. AutoLoss can improve these processes by learning a controller in a data-driven way, and figuring out better update schedules using this controller, as long as the schedule does affect the optimization goal. In this paper, we focus mostly on optimization problems, but note AutoLoss is applicable to alternate processes that involve non-optimization subtasks, such as sampling methods~\citep{griffiths2004finding, ma2015complete}.

\noindent \textbf{Meta learning.} Meta learning~\citep{andrychowicz2016learning, maclaurin2015gradient, wang2016learning, finn2017model, chen2016learning} has drawn considerable interest from the community, and has been recently applied to improve the optimization of ML models~\citep{ravi2016optimization, li2016learning, bello2017neural, fan2018learning}. Among these works, the closest to ours are~\citet{li2016learning, bello2017neural, fan2018learning}.~\citet{li2016learning} propose \emph{learning to optimize} to directly predict the gradient values at each step of SGD. Since the gradients are continuous and usually high-dimensional, directly regressing their values might be difficult, and the learned gradient regressor is nontransferable to new models or tasks. Differently,~\citet{bello2017neural} propose to learn better gradient update rules based on a domain specific language. The learned rules outperform manually designed ones and is generalizable. AutoLoss differs from this line of works -- instead of learning to generate values of updates (gradients), AutoLoss focuses on producing better scheduling of updates. Therefore AutoLoss can model other classes of problems such as scheduling the generator and discriminator training in GANs, or even go beyond optimization  problems. In~\citet{fan2018learning}, a \emph{learning to teach} framework is proposed that a teacher model, trained by optimization metadata, can guide the learning of student models. AutoLoss instantiates the framework in the sense that the teacher model (controller) produces better schedules for the task model (student) optimization.

\noindent \textbf{AutoML.} Also of note is another line of works that apply RL to enable automatic machine learning (AutoML), such as device placement optimization~\citep{mirhoseini2017device}, neural architecture search~\citep{baker2016designing, zoph2016neural}, etc. While addressing different problems, AutoLoss' controller is trained in a similar way~\citep{peters2008reinforcement} for sequential and discrete predictions.

\section{AutoLoss}
\label{sec:autoloss}
\noindent \textbf{Background.} In most ML tasks, given observed data $\mathcal{D}$, we aim to minimize an objective function $\mathcal{L}(\mathcal{D}; \bm{\Theta})$  with respect to the parameters $\bm{\Theta}$ of the model that we use to characterize the data. Solving this minimization problem involves finding the optima of $\bm{\Theta}$ (denoted as $\bm{\Theta}^*$), which we usually resort to a variety of de facto optimization methods~\citep{boyd2004convex} if close-formed solutions are unavailable. In the rest of the paper, we will focus on two typical classes of optimization workflows which many modern ML model solvers would fall into: \emph{iterative} and \emph{alternate} optimization.

Iterative optimization methods look for the optimal parameter $\bm{\Theta}^{*}$ in an \emph{iterative-convergent} way, by repeatedly updating $\bm{\Theta}$ until certain stopping criteria is reached.
Specifically, at iteration $t$, the parameters $\bm{\Theta}$ are updated from $\bm{\Theta}^{(t)}$ to $\bm{\Theta}^{(t+1)}$ following the update equation $\bm{\Theta}^{(t+1)} = \bm{\Theta}^{(t)} + \epsilon \cdot \Delta_{\mathcal{L}}(\mathcal{D}^{(t)}; \bm{\Theta}^{(t)})$, where we denote $\Delta_{\mathcal{L}}$ as the function that calculates update values of $\bm{\Theta}$ depending on $\mathcal{L}$, $\mathcal{D}^{(t)} \subseteq \mathcal{D}$ as a subset of $\mathcal{D}$ used at iteration $t$ and $\epsilon$ a scaled factor. Many widely-adopted algorithms~\citep{bottou2010large,boyd2003subgradient} fall into this family, e.g. in the case for SGD, $\Delta_{\mathcal{L}}$ reduces to deriving the gradient updates $\nabla \bm{\Theta}$ (we skip optional steps such as momentum or projection for clarity), $D^{(t)}$ is a stochastic batch, and $\epsilon$ is the learning rate. 

% TODO (Hao): write a better connection sentence.
To describe alternate optimization,  we notice the objective $\mathcal{L}$ is usually composed of multiple different optimization targets: $\mathcal{L} = \{\ell_m\}_{m=1}^M$, and we want $\bm{\Theta}^*$ to minimize a certain combination of them. For example, when fitting a regression model with mean square error (MSE), appending an L1 loss helps obtain sparsity; in this case, $\mathcal{L}$ is written as a linear combination of MSE and L1 terms. 
Similarly, the parameters $\bm{\Theta}$ in many cases are also composable, e.g. when the model has multiple components with independent sets of parameters. 
If we decompose $\bm{\Theta} = \{\bm{\theta}_n\}_{n=1}^N$, an alternate optimization (in our definition) contains multiple steps, where each step $t$ involves choosing $\ell_{m_t} \in \mathcal{L}, \bm{\theta}_{n_t} \in \bm{\Theta}$, which we will call as \emph{determining an optimization action} (notated as $a$), and update $\bm{\theta}_{n_t}$ w.r.t. $\ell_{m_t}$. 

%set of pairs $\{(\ell, \bm{\theta})| \ell \in \mathcal{L}, \bm{\theta} \in \bm{\Theta}\}$,  and for each pair, updating each $\bm{\theta}$ w.r.t. $\ell$.
%\footnote{While it is also possible to optimize a subset of $\{\mathcal{L}_m\}_{m}^M$ w.r.t. to a subset of $\{\bm{\theta}_n\}_{n=1}^N$, we simplify it for clarity.}. 

Further, we note that many ML optimization tasks in practice are both iterative and alternate, 
%In particular, at each step $t$ during the alternate optimization, we may have no access to a closed form solution for the subproblem ${\min}_{\bm{\theta}} \ell$. We instead update $\bm{\theta}$ iteratively with multiple sub-steps. To see a concrete example, 
%\st{consider finding the equilibrium of a generator $G$ and a discriminator $D$ using SGD in a generative adversarial networks (GAN) framework: one updates the parameters of $G$ and $D$ alternate with multiple steps; while at each step $t$, the gradients of either $G$ or $D$ w.r.t. their specifical optimization target are estimated using a stochastic batch of data, and applied to update its parameters.} 
such as the training process of GANs, where the updates of generator and discriminator parameters are alternated, each with a few iterations of stochastic updates, until equilibrium.

%one iteratively updates the parameters of generator $G$ or discriminator $D$ by applying the gradients of either $G$ or $D$ w.r.t. their optimization target using a stochastic batch of data.
%solving the Latent Dirichlet Allocation (LDA) via variational inference (VI): one usually follows an Expectation-Minimization (EM) framework where the parameters of the model and variational distributions are alternatively optimized in order to maximize the likelihood over training data.
We therefore present iterative and alternate optimization with the following unified formulation:
\begin{equation}
\mbox{for } t = 1 \rightarrow T, \mbox{ choose } (\ell_{m_t}, \bm{\theta}_{n_t}) = a_{q_t} \in \mathcal{A}, \mbox{ update } \bm{\theta}^{(t+1)}_{n_t} = \bm{\theta}^{(t)}_{n_t} + \epsilon \cdot \Delta_{\ell_{m_t}}^{n_t},
%n = 1 \rightarrow N, \quad \bm{\theta}^{(t+1)}_n = \bm{\theta}^{(t)}_n + \epsilon \cdot \sum_{m=1}^M y_{m, n}^{(t)} \cdot \Delta_{\ell_{m}}^n.
\label{eq:unified_workflow}
\end{equation}
where $\mathcal{A} = \{ a_q\}_{q=1}^Q$ denotes the task-specific action space that defines all legitimate pairs of loss and parameter to choose from;  $\Delta_{\ell_{m_t}}^{n_t}$ are update values of ${\bm{\theta}}_{n_t}$ w.r.t. ${\ell}_{m_t}$. Eq.~\ref{eq:unified_workflow} reduces to the vanilla form of iterative optimization when $\mathcal{A} = \{(\mathcal{L}, \bm{\Theta})\}$.

%\begin{equation}
%\small
%\mbox{for } t = 1 \rightarrow T, \quad \bm{\Theta}^{(t+1)}_n = \bm{\Theta}^{(t)}_n + \epsilon \cdot \sum_{m=1}^M y_{\mathcal{\ell}}^{(t)} \cdot \Delta_{\ell_{m}} (\bm{\theta}^{(t)}_n; D^{(t)}),
%\label{eq:unified_workflow}
%\end{equation}
%where we introduce the binary variable $y^{(t)}_{n,m}$ where $y_{\bm{\theta}}^{(t)}(n)$ is $1$  when the $\bm{\theta}_n \in \bm{\Theta}_t$ at the $t$th iteration and $0$ otherwise (similar for $y_{\ell}^{(t)}$),  and $\cup$ as a concatenation operation over multiple vectors. Eq.~\ref{eq:unified_workflow} reduces to the vanilla form of iterative optimization when $y_{\bm{\theta}}^{(t)} (n) = y_{\ell}^{(t)} (m) = 1$, $\forall t, m, n$.

\noindent \textbf{AutoLoss.}
Given the formulation in Eq.~\ref{eq:unified_workflow}, our goal is to determine $a_{q_t}$, i.e. which losses to optimize and what parameters to update at each $t$, in order to maximize the downstream task performance. We introduce a meta model, which we call \emph{controller}, to be distinguished from the \emph{task model} used in the downstream task. The controller is expected to learn during its exploration of task model optimization processes, and is able to decide how to update once sufficient knowledge has been accumulated.

Specifically, we let the controller make sequential decisions at each step $t$; it scans through the past history and the current states of the process (described as a feature vector $\bm{X}^{(t)} \in \mathbb{R}^K$), and predicts a one-hot vector $\bm{Y}^{(t)} \in \{0, 1\}^{|\mathcal{A}|}$, i.e. $a_q \in |\mathcal{A}|$ will be selected if the $q$th entry of $\bm{Y}^{(t)}$ is 1.
%with the entry at $n^$th row and $m^$th column is $y_{n,m}^t$. 
We model our controller as a conditional distribution $p(\bm{y}|\bm{x}; \bm{\phi})$ parameterized by $\bm{\phi}$\footnote{The other alternate is to condition the decision at the $t$ step on the decision made at the $t-1$ step, though we choose a simpler one to highlight the generic idea behind AutoLoss.}, where we denote $\bm{y}$ and $\bm{x}$ as the $|\mathcal{A}|$-dim decision variable and $K$-dim feature variable, respectively. At each step $t$, we sample $\bm{Y}^{(t)} \sim p(\bm{y}|\bm{x} = \bm{X}^{(t)}; \bm{\phi})$, and perform updates following Eq.~\ref{eq:unified_workflow} and $\bm{Y}^{(t)}$. 

\noindent \textbf{Parameter Learning}. 
The parameters of the controller $\bm{\phi}$ is trained to maximize the performance of the optimization task given sampled sequences of decisions within $T$ steps, notated as  $\mathcal{Y} = \{\bm{Y}^{(t)}\}_{t=1}^T$. Accordingly, we introduce the training objective of the controller as $J(\bm{\phi}) = \mathbb{E}_{\mathcal{Y} \sim p(\bm{y}|\bm{x}; \bm{\phi})} \big[ R(\mathcal{Y}) | \mathcal{L}, \bm{\Theta} \big]$, where $R(\cdot)$ is the reward function that evaluates the final task performance after applying the schedule  $\mathcal{Y}$ for its optimization. We will discuss the form of $R$ in \S\ref{sec:applications}.
Since the decision process involves non-differentiable sampling, we learn the parameters using REINFORCE~\citep{williams1992simple} and future variants (see \S\ref{sec:applications} for details)~\citep{schulman2017proximal}, where the unbiased policy gradients at each updating step of the controller are estimated by sampling $S$ sequences of decisions $\{\mathcal{Y}_s\}_{s=1}^S$ (for all experiments we set $S=1$) and compute
\begin{equation}
\nabla_{\phi} J(\phi) = \frac{1}{S} \sum_{s=1}^S \big[ (R(\mathcal{Y}_s) - B) \cdot \nabla_{\phi} \sum_{t=1}^T \log p(\bm{Y}^{(t)}_s |\bm{X}^{(t)}_s; \bm{\phi})| \mathcal{L}, \bm{\Theta} \big],
\label{eq:policy_gradient}
\end{equation}
where $\bm{Y}_s^{(t)}$ is the $t$th decision in $\mathcal{Y}_s$. 
%We next introduce techniques to facilitate the learning.
To reduce the variance, we introduce a baseline term $B$ in Eq.~\ref{eq:policy_gradient} to stabilize the training (similar to ~\citet{pham2018efficient}), where $B$ is defined as a moving average of received reward: $B^{(h+1)} \leftarrow \eta B^{(h)}  + (1-\eta) R^{(h)}$ with $\eta$ as a decay factor. Whenever applicable, the final reward $R(\mathcal{Y}_s) - B$ is clipped to a given range to avoid exploding or vanishing gradients. We present the detailed training algorithm in Appendix~\ref{sec:training_algo}. 

\section{Applications}
\label{sec:applications}
We next apply AutoLoss to four specific ML tasks: $d$-ary quadratic regression and MLP classification with L1 regularization, image generation using GANs, and neural machine translation based on multi-task learning. We instantiate $\mathcal{L}, \bm{\Theta}, \mathcal{A}, \bm{X}^{(t)}$ and $R$ for these tasks.

\subsection{Quadratic Regression and MLP Classification with L1 Regularization}
Given training data $\mathcal{D} = \{\bm{u}_p, v_p\}_{p=1}^P, \bm{u}_{p} \in \mathbb{R}^d, v_p \in \mathbb{R}$ generated by a linear model with Gaussian noise, 
we try to fit them using a $d$-ary quadratic model $f: \mathbb{R}^d \rightarrow \mathbb{R}$ as $f(\bm{u}; \bm{\Theta}) = \bm{u}^{\top} \bm{A} \bm{u} + \bm{b}^{\top} \bm{u} + c$, where parameters $\bm{\Theta} = \{\bm{A}, \bm{b}, c\}$ are optimized via minimizing the mean square error (MSE) $\ell_1(\bm{\Theta}) = \mathbb{E}_{(\bm{u}, v) \in \mathcal{D}} [f(\bm{u; \bm{\Theta}}) - v]^2$. Since fitting the data using a higher-order model is prone to overfitting, we add an L1 term $\ell_{2} = \|\bm{\Theta}\|_1$. A traditional way to find $\bm{\Theta}^{*}$ is to minimize $\ell_1 + \lambda \ell_2$, where $\lambda$ is a hyperparameter yet to be determined by hyperparameter search. This problem can be solved using many iterative optimization methods, e.g. SGD.
To model this problem using AutoLoss, we define $\mathcal{L} = \{\ell_1, \ell_2\}$ ($M=2$), $\bm{\Theta} = \{\bm{\theta}_1\}$ with $\bm{\theta}_1 = \{\bm{A}, \bm{b}, c\}$ ($N=1$), and $\mathcal{A} = \{(\ell_1, \bm{\theta}), (\ell_2, \bm{\theta})\}$, i.e. the controller has Bernoulli outputs which we sample decisions from. 
Similarly, we apply AutoLoss in training a binary MLP classifier $f(\bm{\Theta}): \mathbb{R}^d \rightarrow \{0, 1\}$ with ReLU nonlinearity, which is non-convex and highly prone to overfitting. We materialize $\bm{\Theta} = \{\bm{\theta}_1\}$ where $\bm{\theta}_1$ are all MLP parameters, $\mathcal{L} = \{\ell_1, \ell_2\}$ with $\ell_1$ as the binary cross entropy (BCE) and $\ell_2 = \|\bm{\Theta}\|_{1}$, and $\mathcal{A} = \{(\ell_1, \bm{\theta}), (\ell_2, \bm{\theta})\}$. 

For both tasks, we design $\bm{X}^{(t)}$ as a concatenation of the following features in order to capture the current optimization state and the past history: (1) \emph{training progress}: the percentile progress of training $t/T$. (2) \emph{normalized gradient magnitude}: an $M$-dim vector where the $m$th entry is $\frac{\|\nabla_{\bm{\Theta}} \ell_m\|_2}{\sqrt{\mbox{dim}(\bm{\Theta})}}$. (3) \emph{loss values}: an $M$-dim vector $[\ell_1, \dots, \ell_M]$ that contains values of each $\ell_m$ at $t$. Extracting features (2)(3) requires computing $\ell_m$ and $\nabla_{\bm{\Theta}} \ell_m$ repeatedly at each step $t$, which might be inefficient. We alternatively maintain and use their latest history values -- we compute $\ell_m$ and $\nabla_{\bm{\Theta}} \ell_m$ only when the controller has decided to optimize $\ell_m$ at the jcurrent step, and update their values in the history accordingly. 
(4) \emph{validation metrics}: the loss value of $\ell_1(\bm{\Theta}^{(t)})$ (MSE for regression or BCE for classification) evaluated on a validation set, the exponential moving averages of it and of its higher-order differences. 
Similarly,  we evaluate the validation error only when needed and use their most recent values stored in the history.

%We similarly maintain a history vector and only use the most recent values to avoid repeating evaluating every loss at every step;
For the reward function, we simply instantiated $R = \frac{C}{err}$ for regression and $R = \frac{C}{err - 1}$ for classification, respectively, where $C$ is a constant, $\mbox{err}$ is MSE for regression or classification error for classification, evaluated using converged parameters $\bm{\Theta}^{(T)}$ on the validation dataset. Hence, the controller obtains a larger reward if the task model achieves a lower MSE or classification error.
%We use the same set of features as for $d$-ary regression, except that in this case $\ell_1$ maps to binary cross entropy. We set the reward function as $R = -\frac{C}{1 - \mbox{err}}$ where $\mbox{err}$ is the final classification error evaluated on the validation dataset.%, i.e. the controller is awarded if it learns a distribution of update schedule that 
It is worth noting that we intentionally choose these two models as a proof-of-concept that AutoLoss would work on both convex and non-convex cases. See \S\ref{sec:evaluation} for more experiment results.

\subsection{GANs}
\label{sec:applications:gans}
A vanilla GAN has two set of parameters: the parameters of the generator $G$ as $\bm{\theta}_1$ and those of the discriminator $D$ as $\bm{\theta}_2$, alternately trained via a minimax game as follows (where $\bm{z}$ is notated a noise variable):
\begin{equation*}
\min_{\bm{\theta_1}} \max_{\bm{\theta_2}} \mathcal{L}(\bm{\theta}_1, \bm{\theta}_2) = \mathbb{E}_{\bm{u} \sim p_{data}(\bm{u})} [\log D(\bm{u})] + \mathbb{E}_{\bm{z} \sim p_{\bm{z}}(\bm{z})} [\log (1 - D(G(\bm{z})))].
\end{equation*}
This is a typical alternate process that cannot be expressed by any linear combination of loss terms (hence can hardly benefit from hyperparameter search as in the previous two cases).
How to appropriately balance the optimization of $\bm{\theta}_1$ and $\bm{\theta}_2$ is a key factor that affects the success of GAN training. Beyond fixed schedules, automatically adjusting the training of $G$ and $D$ remains untackled. Fortunately, AutoLoss offers unique opportunities to learn the optimization schedules of GANs.

In particular, we instantiate $\bm{\Theta} = \{\bm{\theta}_1, \bm{\theta}_2\}$, $\mathcal{L} = \{\ell_1, \ell_2\}$ with $\ell_1 = \mathbb{E}_{\bm{z} \sim p_{\bm{z}}(\bm{z})} [\log (1 - D(G(\bm{z})))]$, $\ell_2 = - \mathbb{E}_{\bm{u} \sim p_{data}(\bm{u})} [\log D(\bm{u})] - \mathbb{E}_{\bm{z} \sim p_{\bm{z}}(\bm{z})} [\log (1 - D(G(\bm{z})))]$. To match the possible actions in GANs training, we set $\mathcal{A}$ as $\{(\ell_1, \bm{\theta}_1), (\ell_2, \bm{\theta}_2)\}$, i.e. the controller chooses at each step to optimize one of $G$ and $D$.
To track the training status of both $G$ and $D$, we reuse the same four aspects of features (1)-(4) in previous applications with the following variations: (2) We use a 3D vector $[\frac{\|\nabla_{\bm{\theta}_1} \ell_1\|_2}{\sqrt{\mbox{dim}(\bm{\theta}_1)}}, \frac{\|\nabla_{\bm{\theta}_2} \ell_2\|_2}{\sqrt{\mbox{dim}(\bm{\theta}_2)}}, \log{\frac{\|\nabla_{\bm{\theta}_1} \ell_1\|_2 \cdot \sqrt{\mbox{dim}(\bm{\theta}_2)}}{\|\nabla_{\bm{\theta}_2} \ell_2\|_2 \cdot \sqrt{\mbox{dim}(\bm{\theta}_1)}}}]$, where the first two entries are gradient norms of $G$ and $D$, respectively, while the third is their log ratio to reflect how balanced the updates are; (3) A vector of training losses and their ratio $[\ell_1, \ell_2, \frac{\ell_1}{\ell_2}]$; (4) As there is no clear validation metric to evaluate a GAN, for $G$, we generate a few samples given its current state of parameters $\bm{\theta}^{(t)}_1$, and compute the \emph{inception score} (notated as $\mathcal{IS}$) of them as a feature to indicate how good $G$ is. For $D$, we sample equal number of samples from both $G$ and the training set and use $D$'s classification error (classified as real or fake) on them as a feature. For (2)-(4), we similarly use their most recent history values for improved efficiency. In a same way, we instantiate $R = C \cdot \mathcal{IS}^2$ to encourage the controller to predict schedules that lead to better generators.

\subsection{Multi-task Neural Machine Translation}
Most multi-task learning problems require optimizing several domain-specific objectives jointly for improved performance~\citep{argyriou2007multi}. However, without carefully weighting or scheduling of the optimization of each objective, the results may unexpectedly degrade than optimizing a single objective~\citep{zhang2017survey, teh2017distral}. As the third application, we apply AutoLoss to find better optimization schedules for multi-task learning based neural machine translation (NMT). Following~\citet{niehues2017exploiting}, we build an attention-based encoder-decoder model with three task objectives: the target task translates German into English ($\ell_1$), while the secondary tasks are German named entity recognition (NER) ($\ell_2$) and German POS tagging ($\ell_3$). We use a shared encoder $E$ with parameters $\bm{\theta}_{e}$ and separate decoders $D_{MT}, D_{NER}, D_{POS}$ with parameters as $\bm{\theta}_d^{MT}, \bm{\theta}_d^{NER}, \bm{\theta}_d^{POS}$ for the aforementioned three tasks, respectively. To fit within the AutoLoss framework, we set $\mathcal{L} = \{\ell_m\}_{m=1}^3$, $\bm{\Theta} = \{\bm{\theta}_1, \bm{\theta}_2, \bm{\theta}_3\}$ with $\bm{\theta}_1 = \{\bm{\theta}_e, \bm{\theta}_d^{MT}\}, \bm{\theta}_2 = \{\bm{\theta}_e, \bm{\theta}_d^{NER}\}, \bm{\theta}_3 = \{\bm{\theta}_e, \bm{\theta}_d^{POS}\}$, and the action space $\mathcal{A} = \{(\ell_1, \bm{\theta}_1), (\ell_2, \bm{\theta}_2), (\ell_3, \bm{\theta}_3)\}$, i.e. the controller decides one task to optimize at a time. Still, we reuse the same set of features in previous tasks with small revisions, and set the reward function $R = C \cdot \mbox{PPL}$ where PPL is the validation perplexity. More details about the NMT task are provided in Appendix~\ref{sec:appendix:mtmt_training_detail}.

%A common encoder $E_{ALL}$ is used for all three tasks, but separate decoders $D_{MT}, D_{NER}, D_{POS}$ are used. Parameters of $E_{ALL}, D_{MT}, D_{NER}, D_{POS}$ are denoted as $\bm{\theta_e^{all}}, \bm{\theta_d^{mt}}, \bm{\theta_d^{ner}}, \bm{\theta_d^{pos}}$ We instantiate $\bm{\theta_1} = \{\bm{\theta_e^{all}}, \bm{\theta_d^{mt}}\}$, $\bm{\theta_2} = \{\bm{\theta_e^{all}}, \bm{\theta_d^{ner}}\}$, $\bm{\theta_3} = \{\bm{\theta_e^{all}}, \bm{\theta_d^{pos}}\}$ and $\bm{\Theta} = \{\bm{\theta}_1, \bm{\theta}_2, \bm{\theta}_3\}$ as the set of model parameters, $\mathcal{L} = \{\ell_m\}_{m=1}^3$, and the action space $\mathcal{A} = \{(\ell_1, \bm{\theta}_1), (\ell_2, \bm{\theta}_2), (\ell_3, \bm{\theta}_3)\}$. 

\subsection{Discussion}
When the task model is complex and requires numerous iterations to converge (i.e. when $T$ in Eq.~\ref{eq:policy_gradient} is large), the controller receives sparse and delayed rewards. To facilitate the training, we adapt $T$ depending on the task: for simpler tasks that converge with fewer iterations (e.g. regression and MLP classification), $T$ equals the number of steps to convergence. For GANs and NMT that need longer exploration, we set $T$ as a fixed constant (instead of the max number to convergence) and online train the controller using proximal policy optimization (PPO) algorithm with actor-critic style. We accordingly adjust the reward function as $R^{(t) \rightarrow (t+T)} = C \cdot (P^{(t+T)} - P^{(t)}) / (\frac{P^{(t)} - P^{(t - kT)}}{k})$ where $k$ is a hyperparameter and $P$ is $\mathcal{IS}$ for GANs and $\mbox{PPL}$ for NMT, i.e. we generate a reward every $T$ steps based on the improvement of performance and use it as reward for each step in this segment of steps. Since the improvement will be tiny around optima, we normalize the reward by dividing $\frac{P^{(t)} - P^{(t - kT)}}{k}$ in case the reward is too small to provide enough training signal.

\section{Evaluation}
\label{sec:evaluation}
In this section, we  evaluate AutoLoss empirically on the four tasks using synthetic and real data. We reveal the follow major findings: (1) Overall, AutoLoss can help achieve better quality of convergence faster on all four tasks compared to strong baselines (\S\ref{sec:quality}), with acceptable overheads in controller training. (2) A trained controller on a task model is transferable to guide the training of another task model with different configurations (e.g. neural architectures), or on totally different data distributions, while still converging faster and better (\S\ref{sec:transferability}).
%(3) We investigate the features we used for the controller via ablation studies, and present some empirical insights through AutoLoss' formulation that might be helpful for future model and algorithm development.\footnote{The code of this paper will be publicly available upon acceptance.}

\subsection{Quality of Convergence}
\label{sec:quality}
We first verify the feasibility of the AutoLoss idea. We empirically show that under the formulation of Eq.~\ref{eq:unified_workflow}, there \emph{do exist} learnable update schedules, and AutoLoss is able to capture their distribution and guides the task model to achieve better quality of convergence across multiple tasks and models.

\subsubsection{Regression and Classification with L1 Regularization} \label{sec:regression_classification}
We first apply AutoLoss on two relatively simple tasks with synthetic data, and see whether it can outperform its alternatives (e.g. minimizing linear combinations of loss terms) in combating overfitting. Specifically, for regression, we synthesize dataset $\mathcal{D} = \{\bm{u}_p, v_p\}_{p=1}^P$ using a linear model with Gaussian noise (in the form of $v = \bm{w} \cdot \bm{u} + \xi$ ). In this case, a quadratic regressor is over-expressive and highly likely to overfit the data if without proper regularization. Similarly, for MLP, we synthesize a classification dataset with risks of overfitting by letting only $5\%$ dimensions in $\bm{u}$ be informative whereas the rest be either linear combinations of them or random noise. Details of how the data are synthesized are provided in the Appendix~\ref{sec:appendix:data_synthesis}.
We split our dataset into 5 parts following \citet{fan2018learning}: $\mathcal{D}_{train}^C$ and $\mathcal{D}_{val}^C$ for controller training; Once trained, the controller is used to guide the training of a new task model on another two partitions $\mathcal{D}_{train}^T$, $\mathcal{D}_{val}^T$. Hence, the controller would not work by just memorizing good schedules on $\mathcal{D}_{train}^C$. We reserve the fifth partition $\mathcal{D}_{test}$ to assess the task model after guided training. For both regression and classification, our controller is simply a two-layer MLP with ReLU activation.

\begin{table}
\centering
\begin{tabular}{c|c|c|c|c|c|c}
\hline
Metric & \textsc{w/o L1}  & \textsc{S1}  & \textsc{S2} & \textsc{S3} & \textsc{DGS} & \textsc{AutoLoss} \\
\hline
\hline
\emph{MSE} & .790 & .086 & .096 & .095 & .086 & \bf{.070} \\
\hline
\emph{err} & .124 & .091 & .094 & .094 & .093 & \bf{.088} \\
\hline
\end{tabular}
\caption{\textsc{AutoLoss} vs. \textsc{w/o L1}, schedules S1-S3, and \textsc{DGS} on $d$-ary quadratic regression and MLP classification. Results are averaged over 10 trials. Since the training data are generated with noise, we substitute the baseline MSE (3.94) from the results caused by noise during data generation.}
\vspace{-15pt}
\label{tab:MSE_err}
\end{table}

We compare MSE or classification error (err) evaluated on $\mathcal{D}_{test}$ in Table~\ref{tab:MSE_err} to the following methods: (1) \textsc{w/o L1}: which minimizes only an MSE or BCE term on $\mathcal{D}_{train}^T \cup \mathcal{D}_{val}^T$.
(2) We designed three flexible schedules that optimize the L1 term at each iteration if the condition $\frac{A - B}{B} > th$ is met, where $A, B$ are (S1) task loss values ($\ell_1$) evaluated on $\mathcal{D}_{val}^T$ and $\mathcal{D}_{train}^T$ 
%($$\frac{\ell_1(\mathcal{D}_{val}^T) - \ell_1(\mathcal{D}_{train}^T)}{\ell_1(\mathcal{D}_{train}^T)} > th$)
respectively, (S2) L1 loss and task loss evaluated on $\mathcal{D}_{val}^T$, (S3) gradient norms of L1 and MSE loss. We grid search the threshold $th$ on training data and only report best achieved results.
(3) \textsc{DGS}: we minimize $\ell_1 + \lambda \ell_2$ with $\lambda$ determined by dense grid search (DGS); Particularly, we densely grid search the best $\lambda$ from a pre-selected interval using 50 experiments, and report the best MSE\footnote{Note that the DGS presented is a very strong baseline and might even be unrealistic in practice due to unacceptable cost or lack of prior knowledge on hyperparameters.}.

\begin{wrapfigure}{r}{0.61\textwidth}
\vspace{-18pt}
  \begin{center}
    \includegraphics[width=0.3\textwidth]{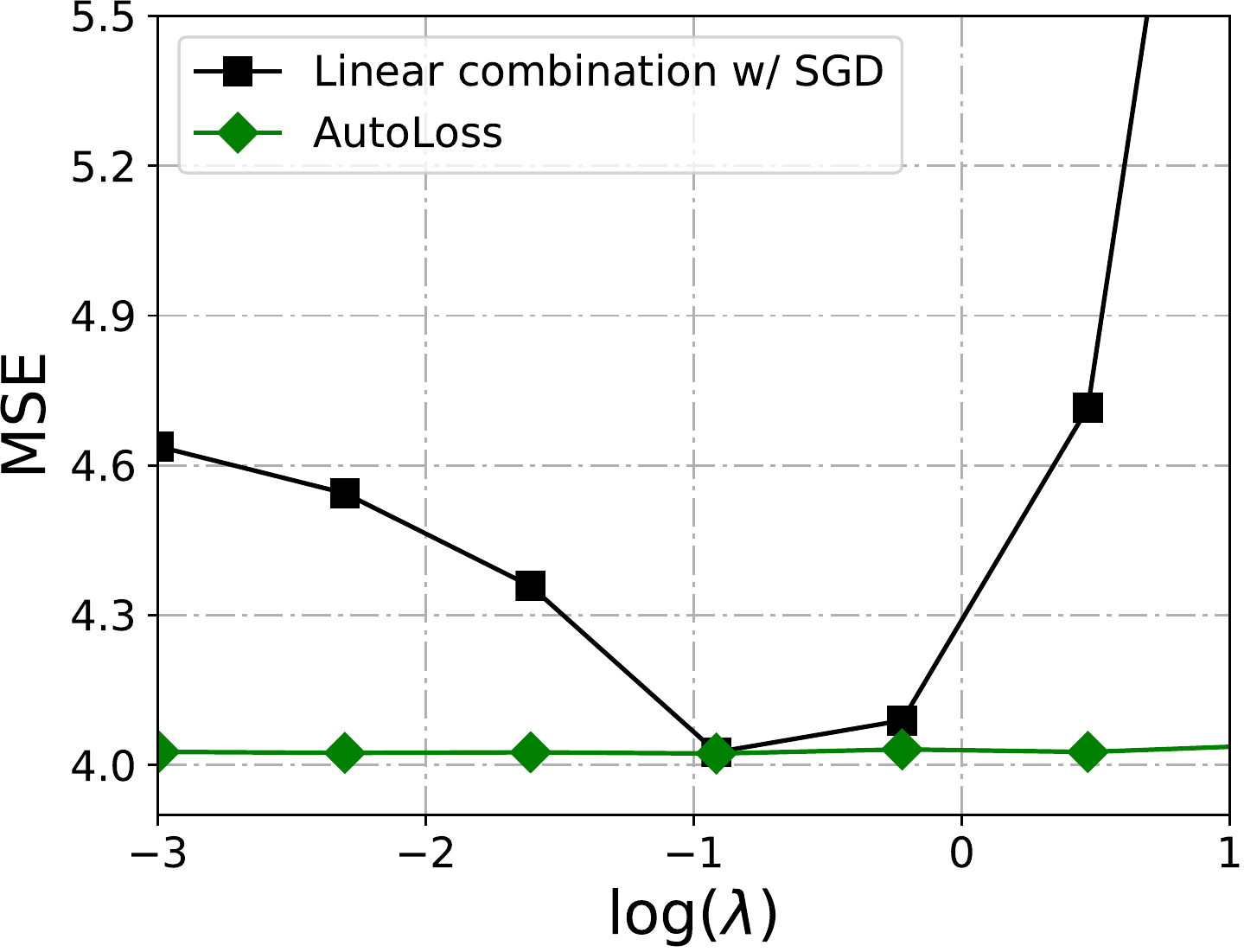}
    \includegraphics[width=0.3\textwidth]{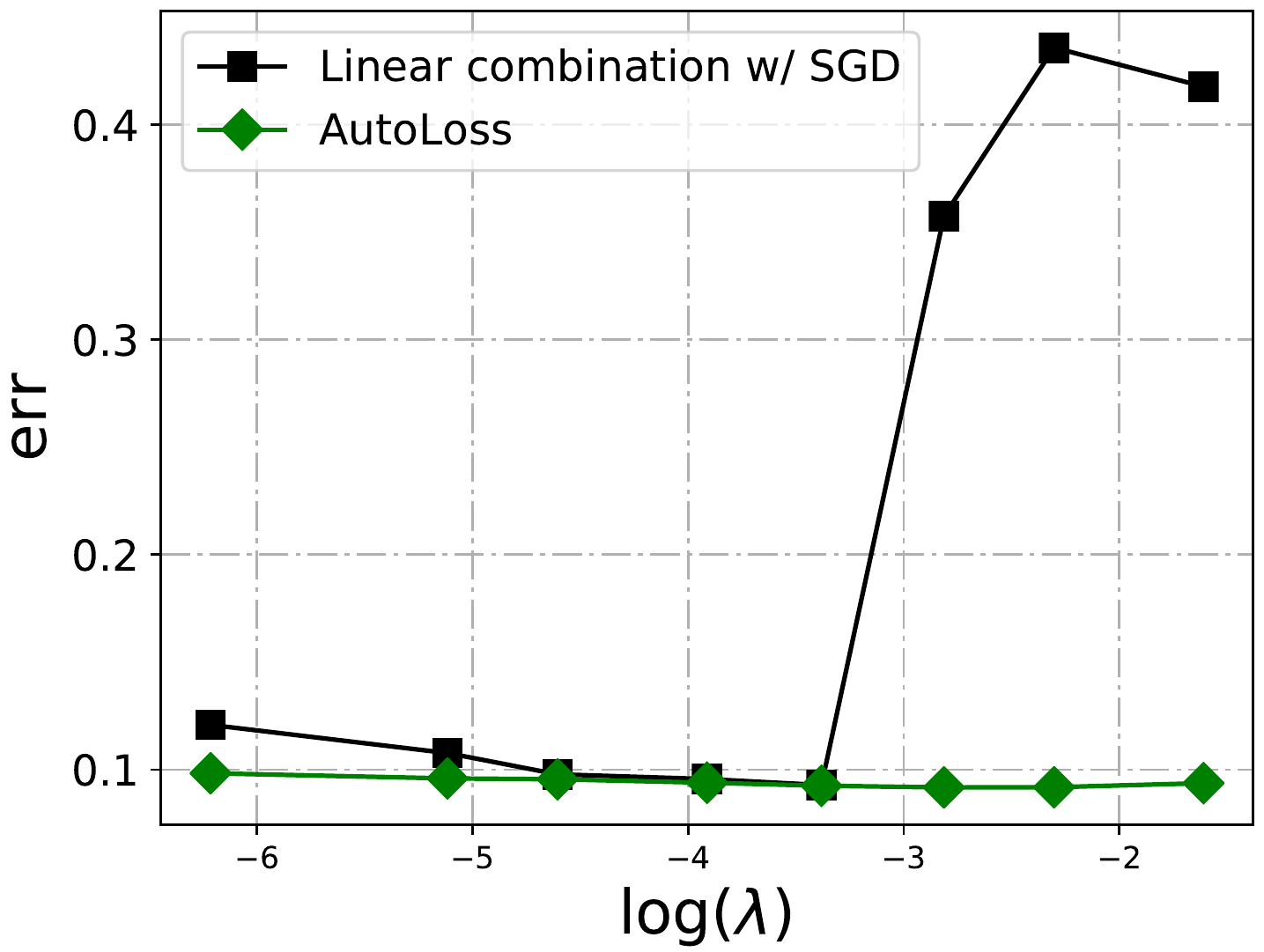}
  \end{center}
  \vspace{-18pt}
 \caption{AutoLoss reaches good convergence regardless of $\lambda$ for both $d$-ary regression (convex) and MLP classification (non-convex).}
 \vspace{-10pt}
\label{fig:regression_lambda}
\end{wrapfigure}
Without regularization, the performance deteriorates -- we observed the large gap between \textsc{w/o L1} and others with L1 on both tasks (convex and non-convex). %MSE on training set is substantially lower than on test set, which, however, is significantly alleviated by adding L1 regularization. 
%This verifies the overfitting risk we created in synthesized data. 
AutoLoss manages to detect and combat the potential risk of overfitting with the designed features, and automatically optimizes the provided L1 term when appropriate. 
In terms of task performance, \textsc{AutoLoss} outperforms three manually designed schedules as well as DGS, a practically very strong method. This is not unexpected as AutoLoss' parametric controller offers more flexibility than heuristic-driven schedules, or any fixed-formed objectives with a dense grid of $\lambda$ values (i.e. DGS). To understand this, consider the $d$-ary quadratic regression which is convex and has global optima only determined by $\lambda$. 
AutoLoss frees the loss surface from being strictly characterized in the form of a linear combination equation, thus allows for finding better optimal solutions that not only enjoy the regularizer effects (i.e. sparsity), but also more closely characterize observed data. As a side benefit, AutoLoss liberates us from hyper-searching $\lambda$, which might be difficult or expensive, and not transferable from one model or dataset to another. 
We perform an additional experiment in Figure~\ref{fig:regression_lambda} where we set different $\lambda$ in $\ell_2 = \lambda |\bm{\Theta}|_2$, and note AutoLoss always reaches the same quality of convergence regardless of $\lambda$. Similar results are observed on MLP classification, a highly non-convex model. The results suggest AutoLoss might be a better alternative to incorporate regularization than fixed-formed combinations of loss terms. We further provide an ablation study on the importance of each designed feature in the Appendix~\ref{sec:appendix:feature_ablation}.

\begin{figure}[tbp]
\centering
\begin{subfigure}{.245\textwidth}
  \centering
  \includegraphics[width=\linewidth]{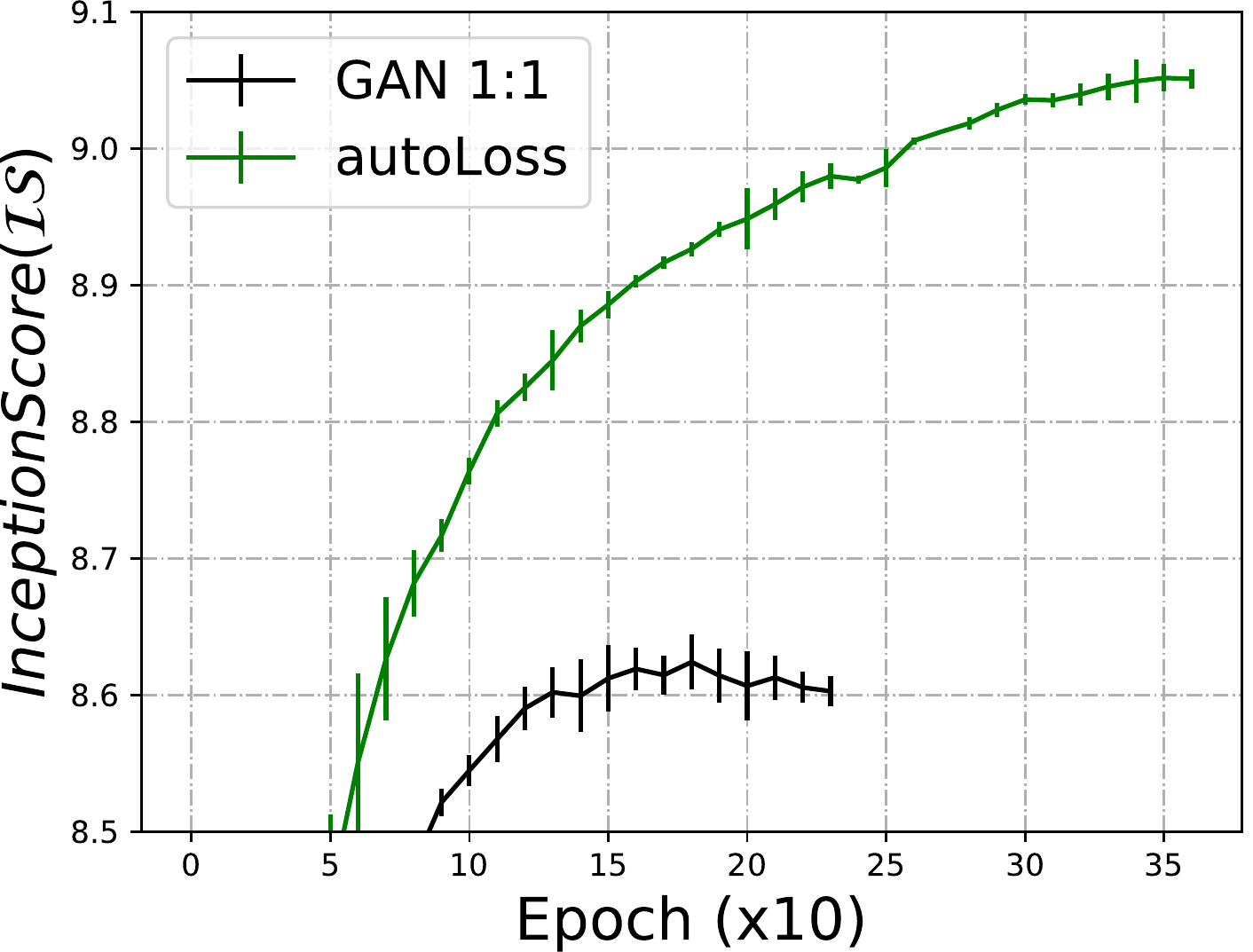}
\end{subfigure}%
\begin{subfigure}{.245\textwidth}
  \centering
  \includegraphics[width=\linewidth]{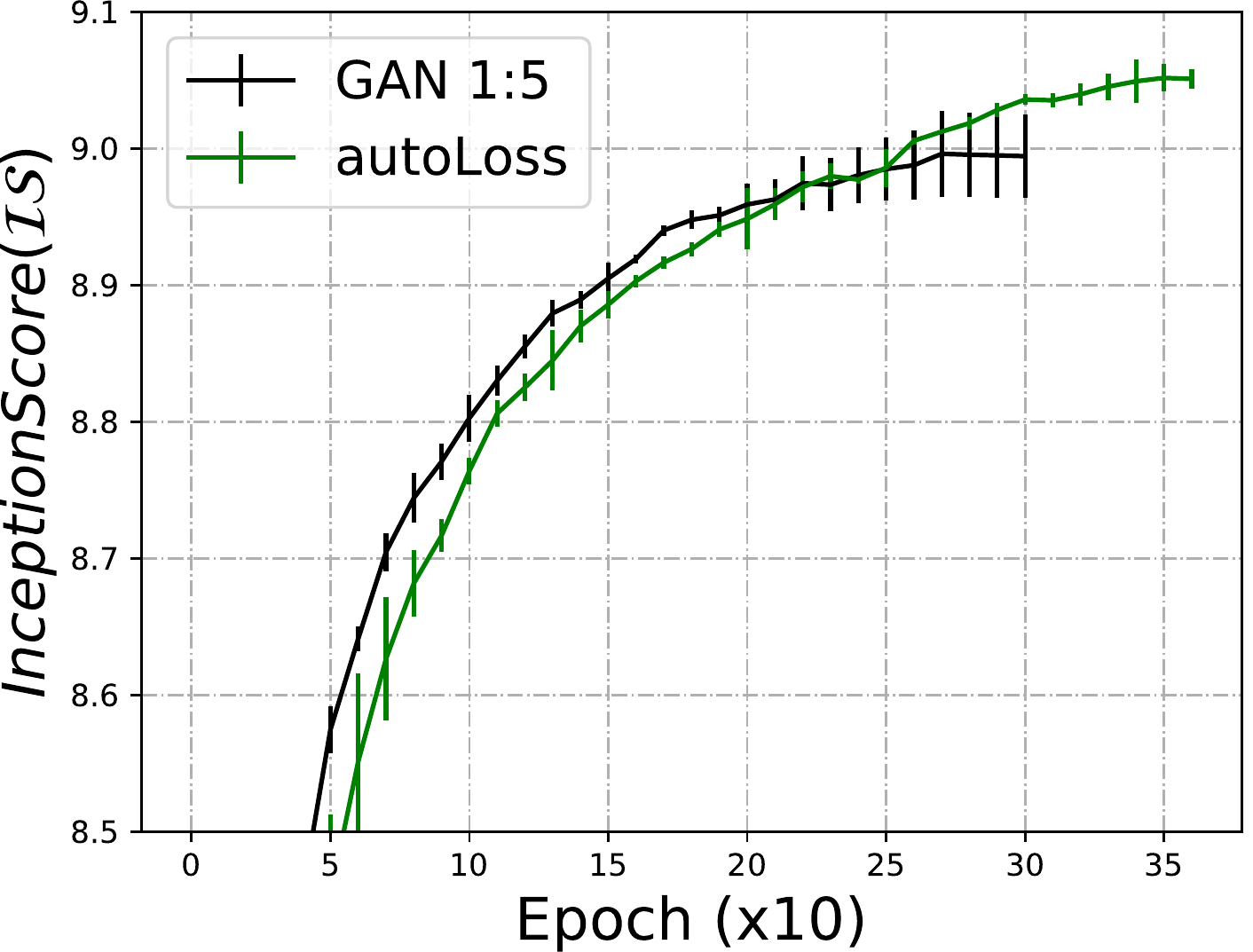}
\end{subfigure}
\begin{subfigure}{.245\textwidth}
  \centering
  \includegraphics[width=\linewidth]{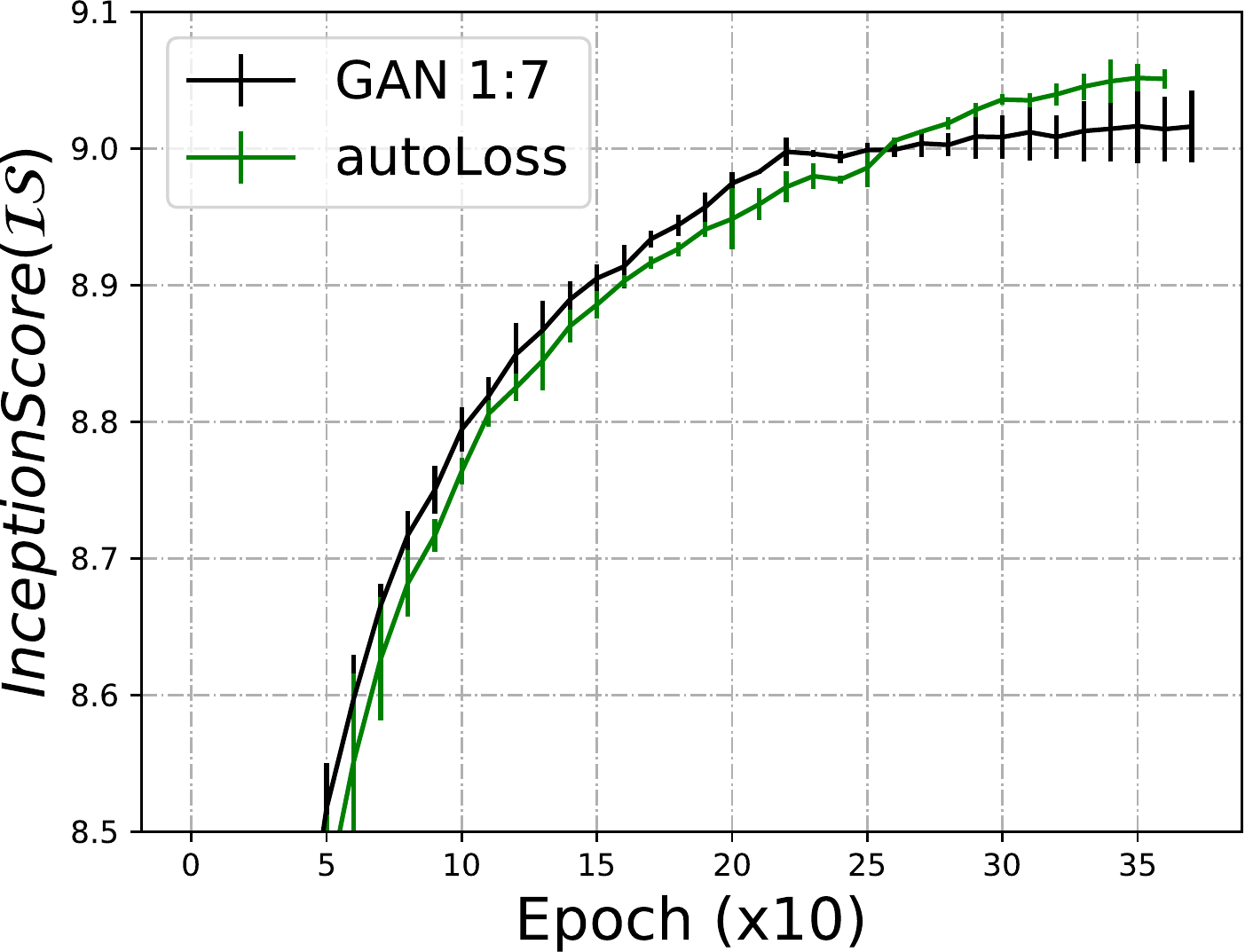}
\end{subfigure}
\begin{subfigure}{.245\textwidth}
  \centering
  \includegraphics[width=\linewidth]{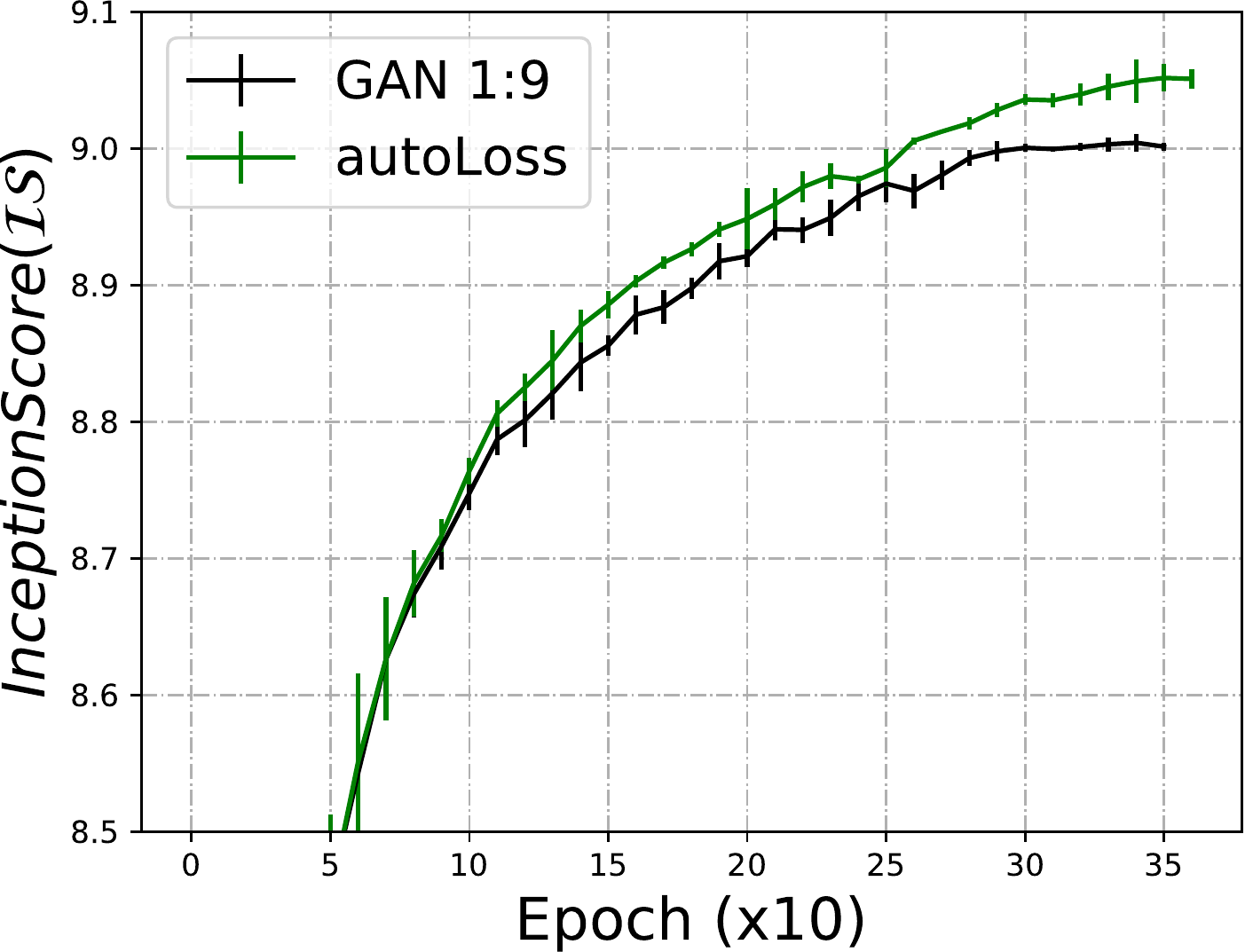}
\end{subfigure}
\vspace{-5pt}
\caption{\small Comparing \textsc{AutoLoss} to 4 best performed baselines out of GAN K:1 and GAN 1:K in terms of training progress ($\mathcal{IS}$ vs. epochs). Each curve corresponds to 3 runs of experiments and the variances are illustrated as vertical lines on the curves.}
\vspace{-15pt}
\label{fig:GAN_progress}
\end{figure}

\subsubsection{GANs}
\label{sec:evaluation:gans}
We next use AutoLoss to help train GANs to generate images. We first build a DCGAN with the architecture of $G$ and $D$ following~\citet{radford2015unsupervised}, and train it on MNIST. As the task model itself is hard to train, in this experiment, we set the controller as a linear model with Bernoulli outputs.
GAN's minimax loss goes beyond the form of linear combinations, and there is no rigorous evidence showing how the training of $G$ and $D$ shall be scheduled. Following common practice, we compare $\textsc{AutoLoss}$ to the following baselines: (1) \textsc{GAN}: the vanilla GAN where $D$ and $G$ are alternately updated once a time; (2) \textsc{GAN 1:K}: suggested by some literature, we build a series of baselines that update $D$ and $G$ at the ratio 1:K (K = 3, 5, 7, 9, 11) in case $D$ is over-trained to reject all samples by $G$; (3) \textsc{GAN K:1}: that we contrarily bias toward more updates for $D$. To evaluate $G$, we use the inception score ($\mathcal{IS}$)~\citep{salimans2016improved} as a quantitative metric, and also visually inspect generated results. To calculate $\mathcal{IS}$ of digit images, we follow~\citet{deng2017structured} and use a trained CNN classifier on MNIST train split as the ``inception network'' (real MNIST images have $\mathcal{IS} = 9.5$ on it). In Figure~\ref{fig:GAN_progress}, we plot the $\mathcal{IS}$
%\footnote{While we are aware of that inception score has recently been doubted~\citep{barratt2018note} as a proper metric to evaluate generators, we visualize the generated samples and observe in our experiments that it is directly relevant with the quality of generated images, therefore could be used as an indicator of the convergence.} 
w.r.t. number of training epochs, comparing \textsc{AutoLoss} to four best performed baselines out of all \textsc{GAN 1:K} and \textsc{GAN K:1}, each with three trials of experiments. We also report the converged $\mathcal{IS}$ for all methods here: 8.6307, 9.0026, 9.0232, 9.0145, \textbf{9.0549} for \textsc{GAN}, \textsc{GAN (1:5)}, \textsc{GAN (1:7)}, \textsc{GAN (1:9)}, \textsc{AutoLoss}, respectively.

In general, GANs trained with AutoLoss present two improvements over baselines: higher quality of final convergence in terms of $\mathcal{IS}$, and faster per-epoch convergence. For example, comparing to \textsc{GAN 1:1}, \textsc{AutoLoss} improves the converged $\mathcal{IS}$ for 0.5, and is almost 3x faster to achieve where \textsc{GAN 1:1} converges ($\mathcal{IS} = 8.6$) in average. We observe \textsc{GAN 1:7} performs closest to \textsc{AutoLoss}: it achieves $\mathcal{IS} = 9.02$, compared to \textsc{AutoLoss} 9.05, though almost 5 epochs slower to converge, and exhibits higher variance in multiple experiments. It is worth noting that all \textsc{GAN K:1} baselines perform worse than the rest and are skipped in Figure~\ref{fig:GAN_progress}, echoing the statements~\citep{arjovsky2017wasserstein,gulrajani2017improved,deng2017structured} that more updates of $G$ than $D$ might be preferable in GAN training. We visualize some generated digit images by AutoLoss-guided GANs in the Appendix~\ref{sec:appendix:MNIST_images} and find the visual quality directly relevant with $\mathcal{IS}$ and no mode collapse is observed.

%Through experiments, we find in the regression task where $\mathcal{L}$ is convex, AutoLoss is able to discover significantly better solutions that may not be within the space of what a linear combination of losses can express; In both the regression and the MLP classifier cases, where the task models are over-expressive, AutoLoss can detect and remedy potential overfitting when necessary; In GANs, AutoLoss is able to learn a parametrized distribution  for update schedule of $G$ and $D$ that achieves both more efficient and higher-quality convergence.  

\begin{figure*}[tbp]
\centering
\begin{subfigure}{.33\textwidth}
  \includegraphics[width=\linewidth]{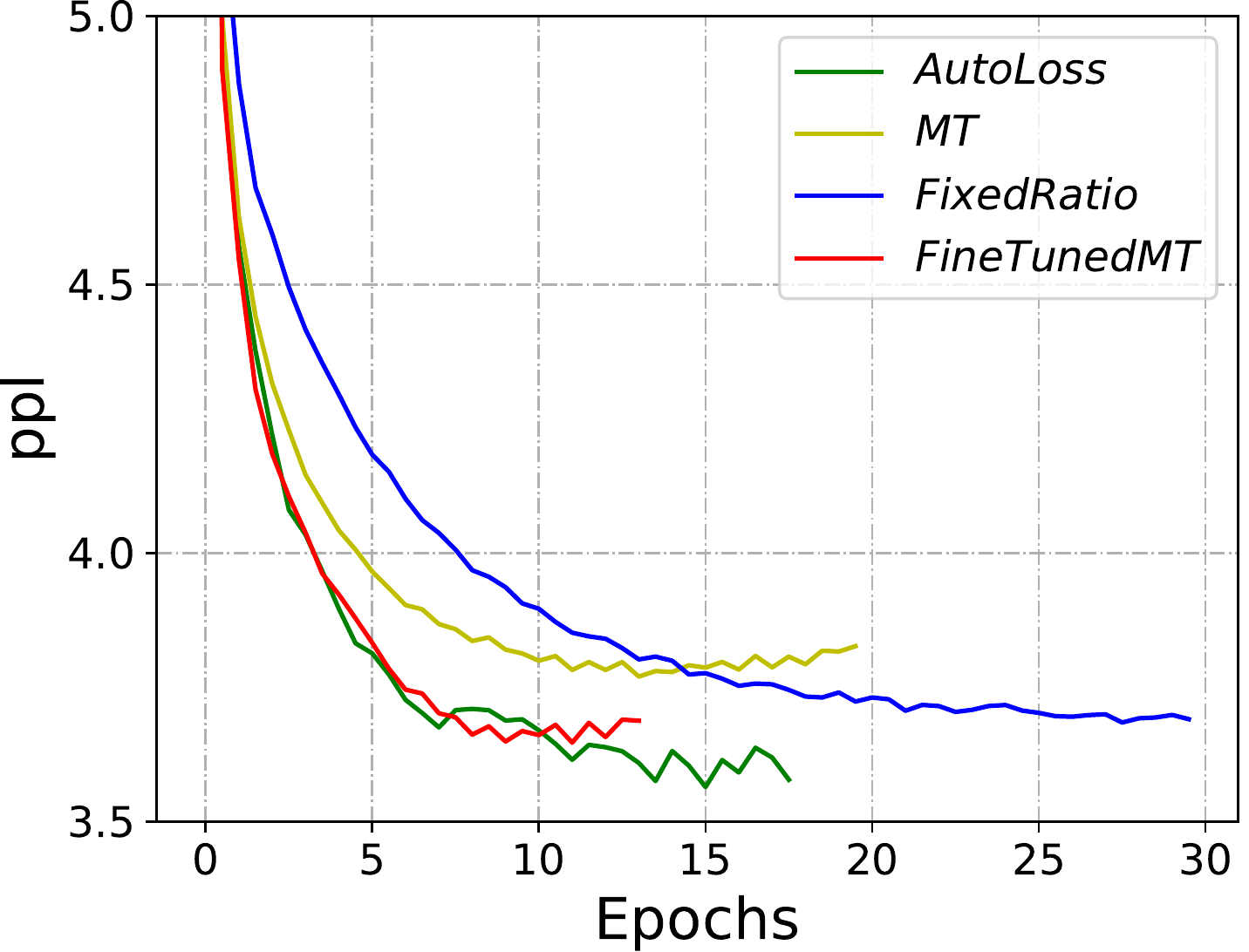}
\end{subfigure}%
\begin{subfigure}{.33\textwidth}
  \includegraphics[width=\linewidth]{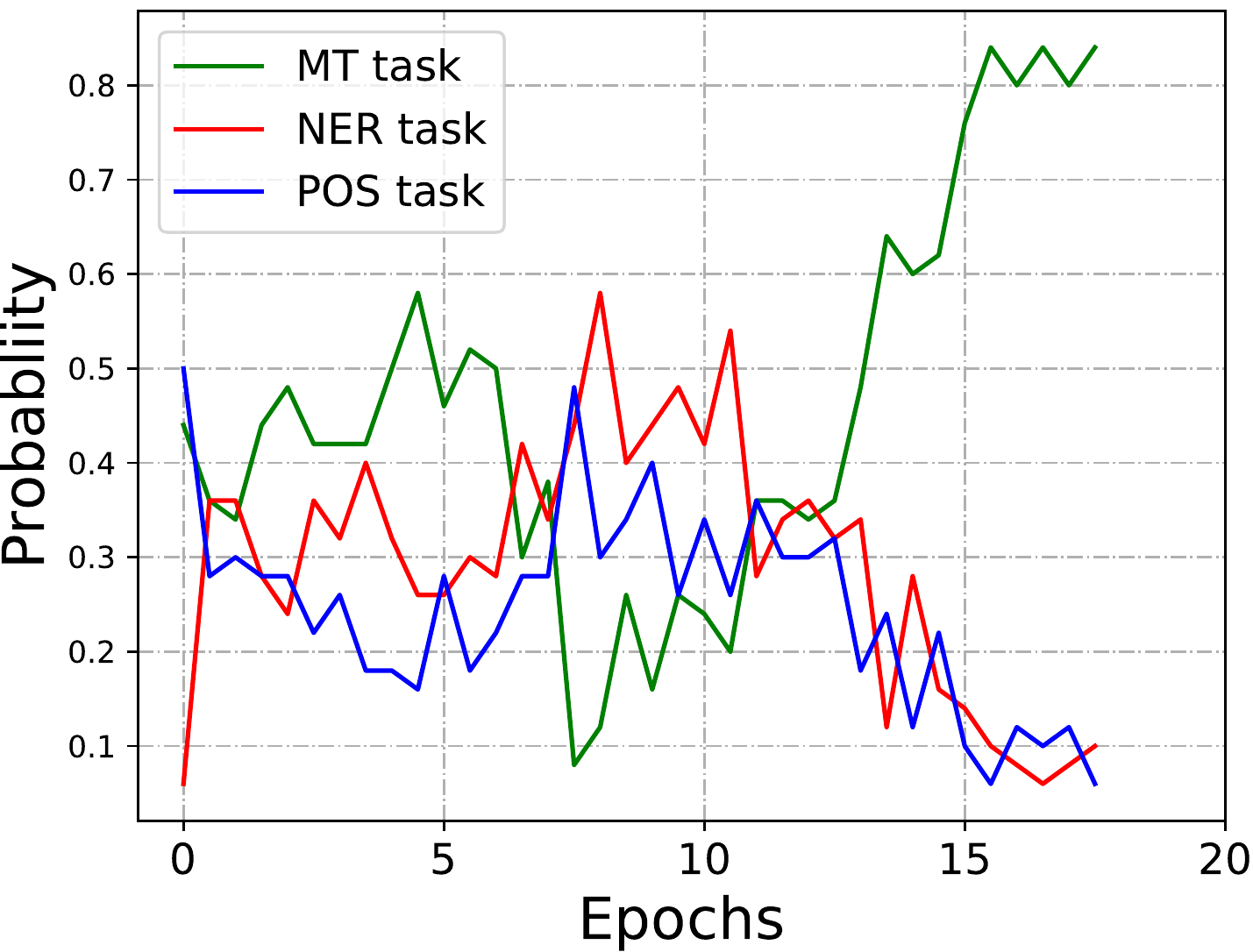}
\end{subfigure}
\begin{subfigure}{.33\textwidth}
  \centering
  \includegraphics[width=\linewidth]{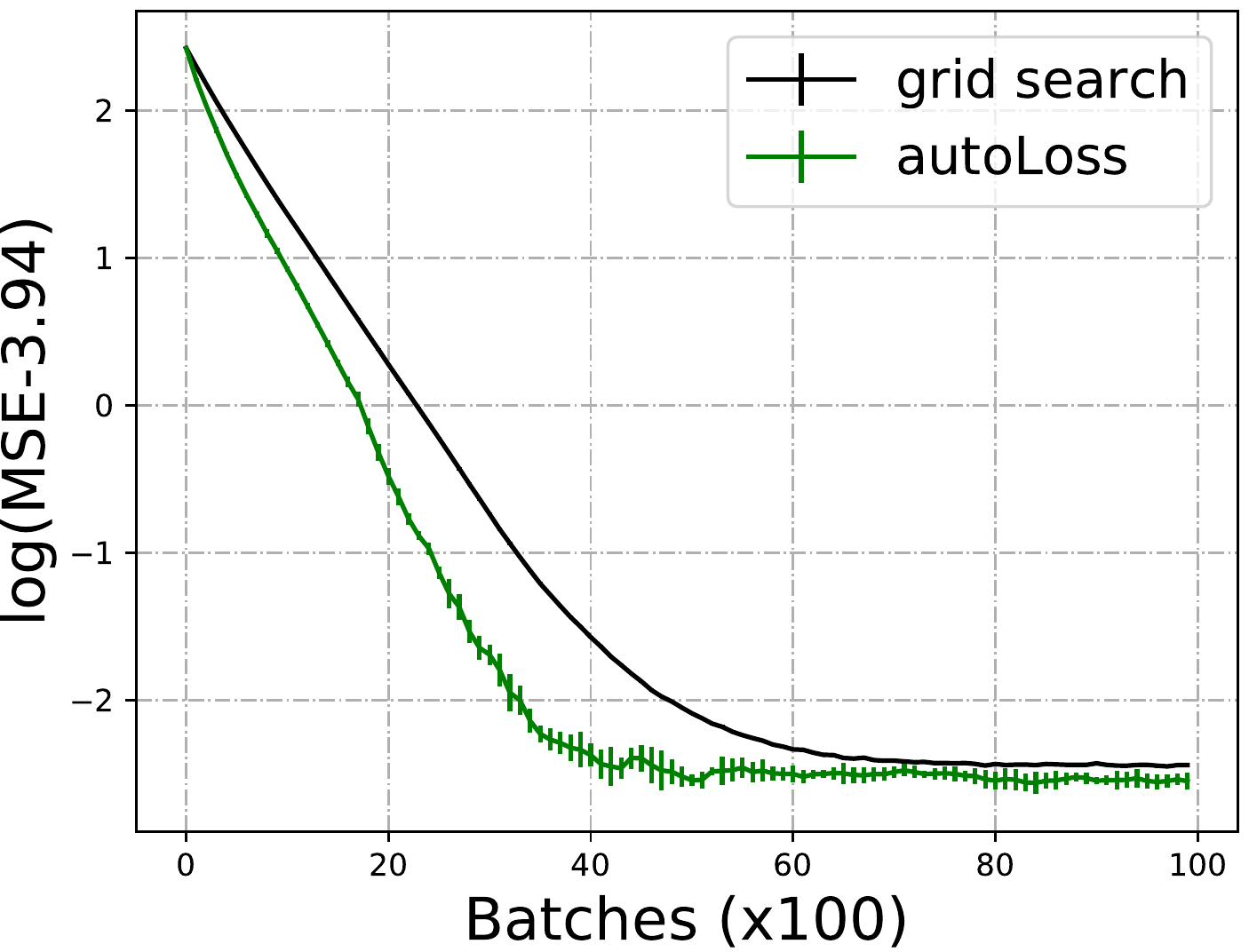}
\end{subfigure}
\vspace{-5pt}
\caption{\small (L) Validaton PPL w.r.t. training epochs on the NMT task; (M) Visualization of the trained controller's policy on the NMT task; (R) \textsc{AutoLoss} vs. \textsc{DGS} in terms of MSE w.r.t. scanned data batches on the $d$-ary regression task.}
\vspace{-10pt}
\label{fig:NTM_budget}
\end{figure*}

\subsubsection{Multi-task Neural Machine Translation}
Lastly, we evaluate AutoLoss on multi-task neural machine translation. Our NN architecture exactly follows the one in~\citet{niehues2017exploiting}. More information about the dataset and experiment settings are provided in Appendix~\ref{sec:appendix:mtmt_training_detail} and~\citet{niehues2017exploiting}. We use an MLP controller with a 3-way softmax output, and train it along with the NMT model training, and compare it to the following approaches: (1) \textsc{MT}: single-task NMT baseline trained with parallel data; (2) \textsc{FixedRatio}: a manually designed schedule that selects which task objective to optimize next based on a ratio proportional to the size of training data for each task; (3) \textsc{FineTuned MT}: train with \textsc{FixedRatio} first and then fine-tune delicately on MT task. Note that baselines (2) and (3) are searched and heavily tuned by authors of~\citet{niehues2017exploiting}. We evaluate the perplexity (PPL) on validation set w.r.t. training epochs in Figure~\ref{fig:NTM_budget}(L), and report the final converged PPL as well: 3.77, 3.68, 3.64, \textbf{3.54} for \textsc{MT}, \textsc{FixedRatio}, \textsc{FineTuned MT} and \textsc{AutoLoss}, respectively. 

We observe that all methods progress similarly but \textsc{AutoLoss} and \textsc{FineTune MT} surpass the other two after several epochs. \textsc{AutoLoss} performs similarly to \textsc{FineTune MT} in terms of training progress before epoch 10, though \textsc{AutoLoss} learns the schedule fully automatically while \textsc{FineTune MT} requires heavy manual crafting. AutoLoss is about 5x faster than \textsc{FixedRatio} to reach where the latter converges, and reports the lowest PPL than all other approaches after convergence, crediting to its flexibility of being able to parameterize and learn the update schedules. We visualize the controller's softmax output after convergence in Fig~\ref{fig:NTM_budget}(M). It is interesting to notice that the controller meta-learns to up-weight the target NMT objective at later phase of the training. This, in some sense, seems to resemble the ``fine-tuning the target task'' strategy appeared in many multi-task learning literature, but is much more flexible thanks to the parametric controller.

\subsection{Overhead}
AutoLoss introduces three possible sources of overheads: controller feature extraction, controller inference and training, and potential cost by additional task model training.
Since we build features merely based on existing metadata or histories (see \S\ref{sec:applications}), which have to be computed anyway even without AutoLoss, the feature extraction has negligible overhead. 
Moreover, as a simple 2-layer MLP controller would suffice for many applications per our experiments, training or inference with the controller add minimal computational overhead, especially on modern hardware such as GPUs. 

Besides, for tasks that converge shortly within a few iterations (e.g. $d$-ary regression and MLP classification), AutoLoss, similar to grid search, requires repeating multiple experiments in order to accumulate sufficient supervisions ($T$ is \# of steps to converge). To assess the resulted overhead, we perform a fixed budget experiment: given a fixed number of data batches allowed to scan, we compare in Fig~\ref{fig:NTM_budget}(R) the reached convergence by \textsc{AutoLoss} and \textsc{DGS} on the regression task. We observe AutoLoss is much more sample-efficient -- it achieves better convergence with less training runs. On the other hand, for computational-heavy tasks that need many steps to converge (GANs, NMT), the controller training, in most cases, can finish simultaneously with task model training, and does not repeat experiments as many times as other hyperparameter search methods would do.

%The cost of AutoLoss controller training usually varies with the task model. On 
%we design an experiment as follows: given a fixed budget of training experiments allowed to run, we compared the best task performance achieved by AutoLoss and DGS, on d-ary quadratic regression and GANs, two tasks with significantly different computational loads. \hao{Can we perform this experiment? besides better quality convergence, AutoLoss controller, once trained, is more transferrable as we will show next.}

%Overhead (1) must be discussed relative to the cost of task model (albeit the absolute overhead is low as a simple MLP controller suffices for many applications): in simple tasks ($d$-ary regression), the controller, similar to grid search, repeats multiple experiments in order to accumulate sufficient supervisions. For computation- and memory- intensive tasks that need many steps to converge (DL, GANs), the controller training needs much less episodes (some can finish together with task model training). The overhead is therefore minimal, e.g. AutoLoss takes 1\% FLOPs of GAN training on MNIST. 

\begin{table}[tbp]
\centering
    \begin{tabular}{c|c|c|c}
    \hline
    Dataset $\#$ & \textsc{w/o L1} & \textsc{DGS} & \textsc{AutoLoss} \\
    \hline
    \hline
    \emph{1} & .1337 & \textbf{.1019} & .1037 \\
    \hline
    \emph{2} & .1294 & .1035  & \textbf{.1016} \\
    \hline
    \emph{3} & .1318 & .1022 & \bf{.0997} \\
    \hline
    \end{tabular}
\vspace{-5pt}
\caption{Comparing \textsc{AutoLoss} to other methods when transferring a trained AutoLoss controller for MLP classification to different data distributions.}
\label{tab:MLP_transfer}
\vspace{-10pt}
\end{table}

\subsection{Transferability}
\label{sec:transferability}
We next investigate the transferability of a trained controller to different models or datasets.

\subsubsection{Transfer to Different Models}
\label{sec:transfer_to_different_models}
\begin{figure}[tbp]
\centering
\begin{subfigure}{0.38\textwidth}
  \includegraphics[width=\linewidth]{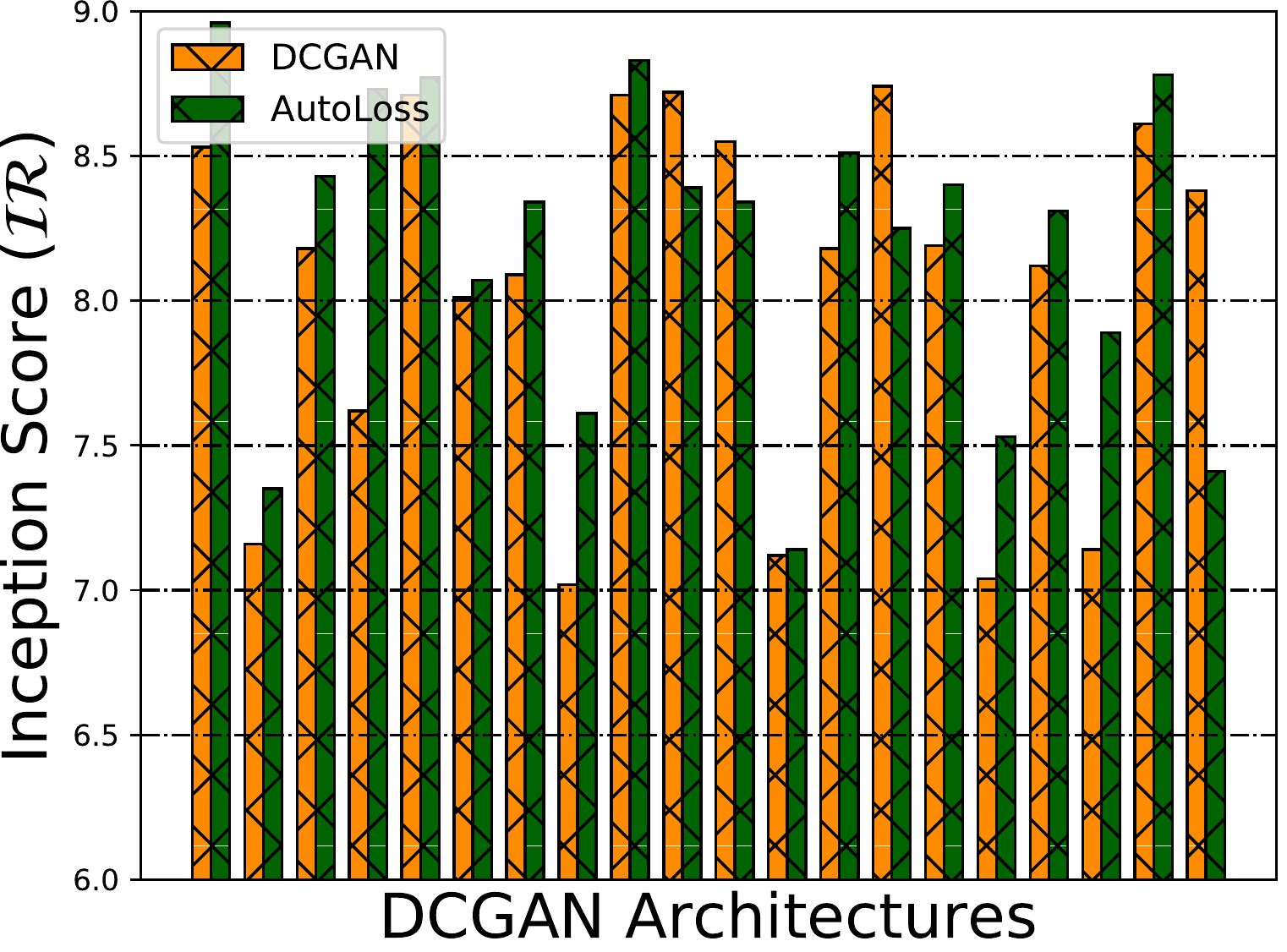}
\end{subfigure}
\begin{subfigure}{0.38\textwidth}
  \includegraphics[width=\linewidth]{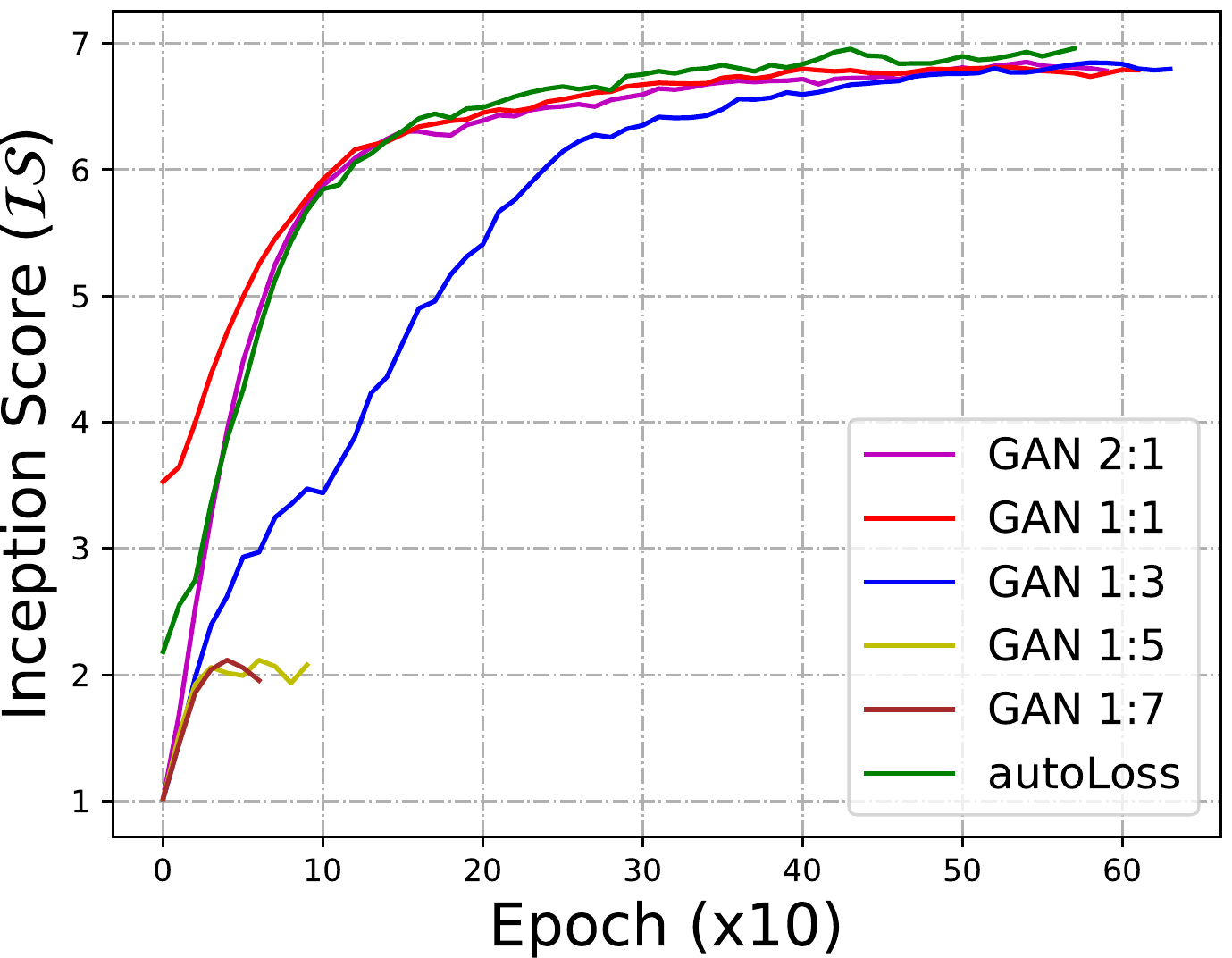}
\vspace{-20pt}
\end{subfigure}
\label{fig:transfer_to_models}
\caption{(L) Comparing the final convergence ($\mathcal{IS}$) on training randomly sampled DCGAN architectures w/ and w/o a pre-trained AutoLoss controller; (R) On CIFAR-10, comparing the training progress ($\mathcal{IS}$ vs. epochs) of a series of baselines GAN 1:K, GAN K:1 and an AutoLoss-guided GAN  with its controller trained on MNIST.}
\vspace{-10pt}
\label{fig:transfer}
\end{figure}

To see whether a differently configured task model can benefit from a trained controller, we design the following experiment: we let a trained DCGAN controller on MNIST guide the training of new GANs (from scratch) whose $G$ and $D$ have randomly sampled neural architectures. We describe the sampling strategies in Appendix~\ref{sec:appendix:DCGAN_sampling}.
%on MNIST; we then change the GAN by randomly sampling the neural architectures of $G$ and $D$, and let the controller guide the training of the new GAN from scratch, on the same dataset.
We compare the (averaged) converged $\mathcal{IS}$ between with and without the AutoLoss controller in Fig~\ref{fig:transfer}(L), while we skip cases that both AutoLoss and the baseline fail ($\mathcal{IS} < 6$) because improper neural architectures are sampled. \textsc{AutoLoss} manages to generalize to unseen architectures, and outperforms \textsc{DCGAN} in 16 out of 20 architectures. This proves that the trained controller is not simply memorizing the optimization behavior of the specific task model it is trained with; instead, the knowledge learned on a neural network is generalizable to novel model architectures.

\subsubsection{Transfer to Different Data Distributions} 
Our second set of experiments try to figure out whether an AutoLoss controller can generalize to different data distributions. Accordingly, we let a trained controller on one dataset to guide the training of the same task model from scratch, but on a different dataset with totally different distributions. We compare the AutoLoss-trained model to other methods, and report the results in Table~\ref{tab:MLP_transfer} and Figure~\ref{fig:transfer}(R) on two tasks respectively: MLP classification, for which we synthesize 4 datasets following a generative process with 4 different specifications (therefore different distributions), with one of them used for controller training; GANs, where we first train a controller for digit generation on MNIST, and use the controller to guide the training of the same GAN architecture on CIFAR-10. In both cases, we observe AutoLoss manages to guide the model training on unseen data. On MLP classification, it delivers trained models comparable to or better than models searched via \textsc{DGS}, while being 50x more economical --- note that DGS has to repeat 50 or more experiments to achieve the reported results in Table~\ref{tab:MLP_transfer} on unseen data (or model). By contrast, AutoLoss, once trained, is free at inference phase.
On image generation, when transferred from digit images to natural images, a controller guided GAN achieves both higher quality of convergence and faster per-epoch convergence than a normal GAN trained with various fixed schedules, among which we observe \textsc{GAN 1:1} performs best on CIFAR-10, while most of \textsc{GAN K:1} schedules fail. 
We visually inspect the images generated by DCGANs guided by the MNIST-trained controller and find the image quality satisfying and no mode collapse occurred, with converged $\mathcal{IS} \approx 7$, compared to best reported $\mathcal{IS} = 6.16$ by DCGANs in previous literature. Visualization of the generated CIFAR-10 images can be found in Appendix~\ref{sec:appendix:CIFAR10_images}.

Finally, we are also interested in knowing whether a trained controller is transferable when both data and models change. We transfer a DCGAN controller trained on MNIST to a new DCGAN with different architectures on CIFAR-10, and observe comparable quality and speed of convergence to the best fixed schedule on CIFAR-10, though AutoLoss bypasses the schedule search and is more readily available.

\section{Conclusion}
We propose a unified formulation for iterative alternate optimization and developed AutoLoss, a framework to automatically learn and generate optimization schedules. Comprehensive experiments on synthetic and real data have demonstrated that the optimization schedule produced by AutoLoss controller can guide the task model to achieve better quality of convergence, and the trained AutoLoss controller is transferable from one dataset to another, or one model to another.

\newpage
\bibliography{autoloss}
\bibliographystyle{abbrvnat}

\newpage
\appendix
\section{Appendix}
\subsection{Training Algorithm}
\label{sec:training_algo}
In addition to the descriptions in \S\ref{sec:autoloss}, we present the detailed training procedures in Algorithm~\ref{algo:training}. 
For all our experiments, we set $S = 1$. For simple tasks such as d-ary regression and MLP classification that converge quickly in a few steps (therefore less costly), we set $T$ as the number of iterations took for a training instance to converge, i.e. a reward is generated upon the completion of a training instance, and we repeat multiple training instances until the controller has converged. For computational-heavy tasks such as GANs that require many iterations to converge, we set $T$ as a fixed constant, meaning that we evaluate $R$ to generate an intermediate reward every $T$ steps (before convergence), and perform a policy gradient update step, in case the exploration takes too long and the reward is too sparse.

\begin{algorithm}[htpb]
\begin{algorithmic}[1]
\small
\State Determine task-specific parameters $S$ and $T$.
\Repeat
\For{$s = 1 \rightarrow S$}
\For{$t = 1 \rightarrow T$}
\State Extracting the optimization state feature $\bm{X}^{(t)}$.
\State Determine the optimization action $a_{q_t} = (\ell_{m_t}, \bm{\theta}_{n_t})$ by sampling $\bm{Y}^{(t)} \sim p(\bm{y} | \bm{x} = \bm{X}^{(t)}; \bm{\phi})$.
\State Perform one step of the task model optimization:  $\bm{\theta}_{n_t} \leftarrow \bm{\theta}_{n_t} + \epsilon \cdot \Delta_{\ell_{m_t}}^{n_t}$
\EndFor
\State Evaluate the reward $R(\mathcal{Y}_s) = R(\{\bm{Y}^{(t)}\}_{t=1}^T)$ received so far, and generate a pair $(\mathcal{Y}_s, R(\mathcal{Y}_s))$.
\EndFor
\State Update controller parameters $\phi$ using Eq.~\ref{eq:policy_gradient} and $\{(\mathcal{Y}_s, R(\mathcal{Y}_s))\}_{s=1}^S$
\Until{convergence.}
\end{algorithmic}
\caption{Training AutoLoss controller along with a task model.}
\label{algo:training}
\end{algorithm}

%\subsection{Feature Extraction based on history}
%Most of the features presented in \S\ref{sec:applications} can be extracted in a highly efficient way by maintaining a history of their values and using only their latest version in the history. Empirically we find using history-based features significantly improves computational efficiency while only slightly compromises the performance.

\subsection{Data Synthesis for $d$-ary Quadratic Regression and MLP Classification}
\label{sec:appendix:data_synthesis}
For the experiments in \S\ref{sec:regression_classification}, we generate the dataset $\mathcal{D} = \{\bm{u}_p, v_p\}_{p=1}^P$ for the $d$-ary quadratic regression task as follows:
\begin{itemize}
    \item Sample the weight vector $\bm{w} \sim \mbox{Uniform}[-0.5, 0.5]$.
    \item For $p = 1 \rightarrow P$:
    \begin{itemize}
        \item Sample the feature vector $\bm{u}_p \sim \mbox{Uniform}[-5, 5]$.
        \item Sample a Gaussian noise $\xi_p \in \mathcal{N}(0, 2)$.
        \item Generate $v_p = \bm{w}^T \cdot \bm{u}_p + \xi_p$.
    \end{itemize}
\end{itemize}
In our experiments, the synthesized dataset has a groundtruth MSE 3.94 (becasue of the Gaussian noise introduced), which we have subtracted from our results.

For the MLP classification task, we synthesize the data $\mathcal{D} = \{\bm{u}_p, v_p\}_{p=1}^P$ as follows. 
\begin{itemize}
    \item Create four cluster centers $\{\mathcal{C}^c\}_{c=1}^{4}$  by sampling from the vertices of a hypercube.
    \item Assign two centers $\mathcal{C}^1, \mathcal{C}^2$ as positive ($v = 1$) while $\mathcal{C}^3, \mathcal{C}^4$ as negative ($v=0$).
    \item For $p = 1 \rightarrow P$:
    \begin{itemize}
        \item Sample the label $v_p \sim \{0, 1\}$.
        \item Sample $\mathcal{C}$ from $\{\mathcal{C}^1, \mathcal{C}^2\}$ if $v_p = 1$ otherwise from $\{\mathcal{C}^3, \mathcal{C}^4\}$.
        \item Sample $\xi_p \sim \mathcal{N}(0, 1)$ and generate a vector $\bm{u}_p^1 = \mathcal{C} + \xi_p$ as the first 5\% dimensions of $\bm{u}_p$.
        \item Generate $\bm{u}_p^2$ as another 5\% dimensions of $\bm{u}_p$, by randomly linearly combining the dimensions in $\bm{u}_p^1$,
        \item Generate $\bm{u}_p^3$ as the rest dimensions of $\bm{u}_p$, by sampling from $\mathcal{N}(0, 1)$.
        \item Generate $\bm{u}_p = [\bm{u}_p^1, \bm{u}_p^2, \bm{u}_p^3]$
    \end{itemize}
\end{itemize}
 
%Each data point is composed of two informative features $\bm{v_i}$, two redundant features $\bm{v_r}$ and 28 arbitrary noise features $\bm{v_n}$. The informative features are sampled as follows: First we initially creates four cluster centers $\{\mathcal{C}^c\}_{c=1}^{4}$ each located around the vertices of a 2D hypercube with sides of 2 and assign two clusters to each class. For each cluster, we independently sample two features $\bm{u}$ from $\mathcal{N}(0, 1)$. Then we placed the clusters on the vertices of the hypercube to get the informative features, i.e. $\bm{v^c_i}=\mathcal{C}^c + \bm{u}$. The redundant features are random linear combinations of the informative features. Each arbitrary noise feature is independently sampled from $\mathcal{N}(0, 1)$. The overall features of a sample is a concatenation of $\bm{v_i}$, $\bm{v_r}$ and $\bm{v_n}$.
\subsection{Details for Multi-task Neural Machine Translation}
\label{sec:appendix:mtmt_training_detail}
\subsubsection{Data}
For the machine translation task, we use the \textsc{WIT} corpus \citep{cettoloEtAl:EAMT2012} for German to English translation. To accelerate training, we only use one fourth of all data, which has 1M tokens. For the POS tagging task, we use the Tiger Corpus \citep{Brants2004}. The POS tag set consists of 54 tags. The German named-entity tagger is trained on GermEval 2014 NER Shared Task data \citep{benikova2014germeval}. The corpus is extracted from Wikipedia with the the tag set consisting of 24 tags.

We preprocess the data by tokenizing, true-casing and replacing all Arabic number by zero. In addition, we apply byte-pair encoding with 10K subwords on source and target side of the WIT corpus separately. We then apply the subwords to all German and English corpora.

\subsubsection{Architecture}
For the task model, we use an attentional encoder-decoder architecture. The three tasks share one encoder $E$ but have their own decoders $D_{MT}, D_{NER}, D_{POS}$. The encoder is a two-layer bidirectional LSTM with 256 hidden units. All decoders are also two-layer bidirectional LSTMs with luong attention \citep{DBLP:journals/corr/LuongPM15} on the top layer. All hidden sizes in decoders are 256. The word embeddings have a size of 128. 

For the controller model, instead of REINFORCE, we apply Proximal Policy Optimization algorithm (PPO) \citep{DBLP:journals/corr/SchulmanWDRK17} to train the controller. Both actor net and critic net are two-layer MLPs with hidden size 32. Discount rate $\gamma$ is set to 0.95.

\subsubsection{Training}
For the task model, we use Adam optimizer with learning rate 0.0005. Dropout rate is 0.3 at each layer. All gradients are clipped to 1. Batch size is 128. For the controller model, we use Adam optimizer with learning rate 0.001. Buffer size is 2000, batch size is 64. Target policy net and behavior policy net are synchronized every 10 steps of updating.

\subsection{Feature Ablation Study}
\label{sec:appendix:feature_ablation}
We investigate the importance of the designed controller features presented in \S\ref{sec:applications}. In particular, we report in Table~\ref{tab:feature_importance} the performance on the regression task after dropping one of the features, where we find that all features being useful while feature (3), which captures the most recent values of all loss terms, bringing the biggest improvement.
We also tried current and historical states of parameters, gradients, and momentums, and found the set of features presented in \S\ref{sec:applications} achieve best trade-off on performance and efficiency.

\begin{table}
\centering
\begin{tabular}{c|c}
\hline
Feature to drop & \texttt{MSE} \\
\hline
\hline
(2) normalized gradient magnitude & .086 \\
\hline
(3) loss values& .101 \\
\hline
(4) validation metrics & .085 \\
\hline
None & \textbf{.070} \\
\hline
\end{tabular}
\caption{The MSE performance on the regression task when some features presented in \S\ref{sec:applications} are ablated.}
\label{tab:feature_importance}
\end{table} 

\begin{figure}[tbp]
\centering
  \includegraphics[width=0.49\linewidth]{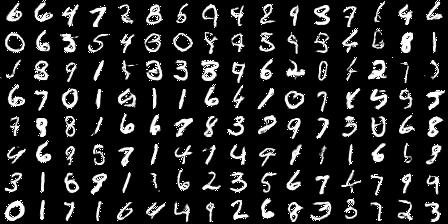}
  \includegraphics[width=0.49\linewidth]{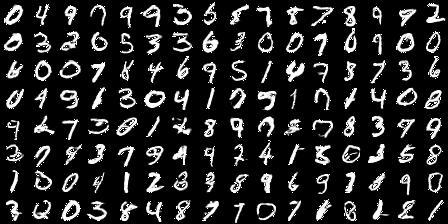}
\caption{\small Images generated by a DCGAN trained under AutoLoss' controller on MNIST. The controller is trained along with the DCGAN training.}
\label{fig:mnist_images}
\end{figure}
\begin{figure}[tph]
\centering
  \centering
  \includegraphics[width=0.48\linewidth]{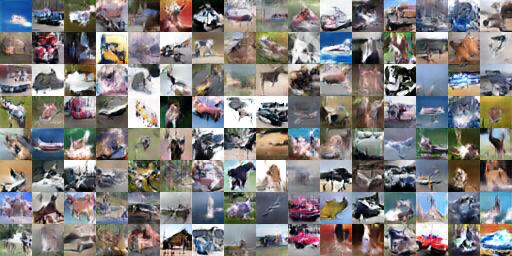}
  \includegraphics[width=0.48\linewidth]{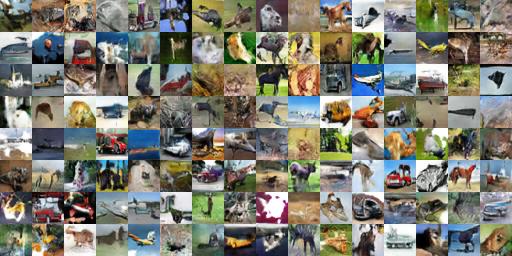}
\caption{\small Images generated by an AutoLoss-guided DCGAN. The controller is trained on MNIST dataset and applied to guide the training of GANs on CIFAR-10. }
\label{fig:cifar10_images}
\end{figure}

\subsection{Sample Strategies to Generate Random DCGAN Architectures}
\label{sec:appendix:DCGAN_sampling}
For the experiments in \S\ref{sec:transfer_to_different_models}, we generate DCGAN architectures~\citep{radford2015unsupervised,salimans2016improved} by randomly sampling the following configurations: 
\begin{itemize}
    \item Sample the number of filters in the base layer of $G$ and $D$ from $\{32, 64, 128\}$.
    \item Sample $\mbox{dim}(\bm{z})$ from $\{64, 128\}$.
    \item Decide whether to use batchnorm or not.
    \item Sample the activation functions from $\{\mbox{ReLU}, \mbox{LeakyReLU}\}$.
\end{itemize}
This results in $3 \times 2 \times 2 \times 2 = 24$ possible DCGAN architectures, among which some of them fail to converge during its training according to our experiments.

\subsection{Image Generated by AutoLoss-guided GANs on MNIST}
\label{sec:appendix:MNIST_images}
In addition to \S\ref{sec:evaluation:gans}, we illustrate in Fig~\ref{fig:mnist_images} the digit images generated by DCGANs trained under AutoLoss' policy, which report the highest $\mathcal{IS} = 9.0549$ on MNIST than any other fixed schedule.

\subsection{CIFAR-10 Images Generated by GANs guided by an AutoLoss Controller Trained on MNIST}
\label{sec:appendix:CIFAR10_images}
In Fig~\ref{fig:cifar10_images}, we illustrate some images generated by DCGANs under guided training of an AutoLoss controller trained on MNIST (with $\mathcal{IS} \approx 7$). We observe the visual quality of generated images are reasonably good.

\end{document}